%% file: paper_template.tex
\documentclass[conference]{IEEEtran}

\usepackage{times}

\usepackage[numbers]{natbib}
\usepackage{multicol}
\usepackage{tabularx}
\usepackage{amsmath,amssymb}
\usepackage{wrapfig}
\usepackage{url}
\usepackage{graphicx}
\usepackage{subcaption}
\usepackage{booktabs, makecell, multirow}
\usepackage[bookmarks=true]{hyperref}
\usepackage{stfloats}
\usepackage[table]{xcolor}
\usepackage{stfloats}

\begin{document}

\title{StereoAdapter-2: Globally Structure-Consistent Underwater Stereo Depth Estimation}

\author{
    \textbf{Zeyu Ren}$^{1*}$\quad \textbf{Xiang Li}$^{2*}$\quad \textbf{Zeyu Zhang}$^{2*\dag}$\quad \textbf{Hao Tang}$^{2\ddag}$ \vspace{0.1cm}\\
    $^1$The University of Melbourne\quad
    $^2$Peking University\\
    \footnotesize $^*$Equal contribution. $^\dag$Project lead.
    $^\ddag$Corresponding author: bjdxtanghao@gmail.com.
}

\maketitle

\begin{abstract}
Stereo depth estimation is fundamental to underwater robotic perception, yet suffers from severe domain shifts caused by wavelength-dependent light attenuation, scattering, and refraction. Recent approaches leverage monocular foundation models with GRU-based iterative refinement for underwater adaptation; however, the sequential gating and local convolutional kernels in GRUs necessitate multiple iterations for long-range disparity propagation, limiting performance in large-disparity and textureless underwater regions. In this paper, we propose StereoAdapter-2, which replaces the conventional ConvGRU updater with a novel ConvSS2D operator based on selective state space models. The proposed operator employs a four-directional scanning strategy that naturally aligns with epipolar geometry while capturing vertical structural consistency, enabling efficient long-range spatial propagation within a single update step at linear computational complexity. Furthermore, we construct UW-StereoDepth-80K, a large-scale synthetic underwater stereo dataset featuring diverse baselines, attenuation coefficients, and scattering parameters through a two-stage generative pipeline combining semantic-aware style transfer and geometry-consistent novel view synthesis. Combined with dynamic LoRA adaptation inherited from StereoAdapter, our framework achieves state-of-the-art zero-shot performance on underwater benchmarks with 17\% improvement on TartanAir-UW and 7.2\% improvement on SQUID, with real-world validation on the BlueROV2 platform demonstrates the robustness of our approach.
Code: \url{https://github.com/AIGeeksGroup/StereoAdapter-2}.
Website: \url{https://aigeeksgroup.github.io/StereoAdapter-2}.
\end{abstract}

\section{Introduction}

Stereo depth estimation serves as a cornerstone for robotic perception, providing metric 3D reconstruction from passive binocular cameras that underpins autonomous navigation~\cite{tartanair}, manipulation, and environmental mapping. In underwater domains, accurate depth sensing is indispensable for AUV/ROV operations spanning infrastructure inspection, ecological monitoring, and archaeological survey, where geometric fidelity directly governs mission safety and autonomy~\cite{revised_underwater}. Nevertheless, underwater imaging introduces pronounced domain shifts stemming from wavelength-dependent attenuation, forward and backscattering, and refraction at water–glass interfaces, which severely violate the photometric consistency assumptions underlying terrestrial stereo pipelines~\cite{uwstereo,UWNet}.

\begin{figure}[t]
  \centering
  \includegraphics[width=\columnwidth]{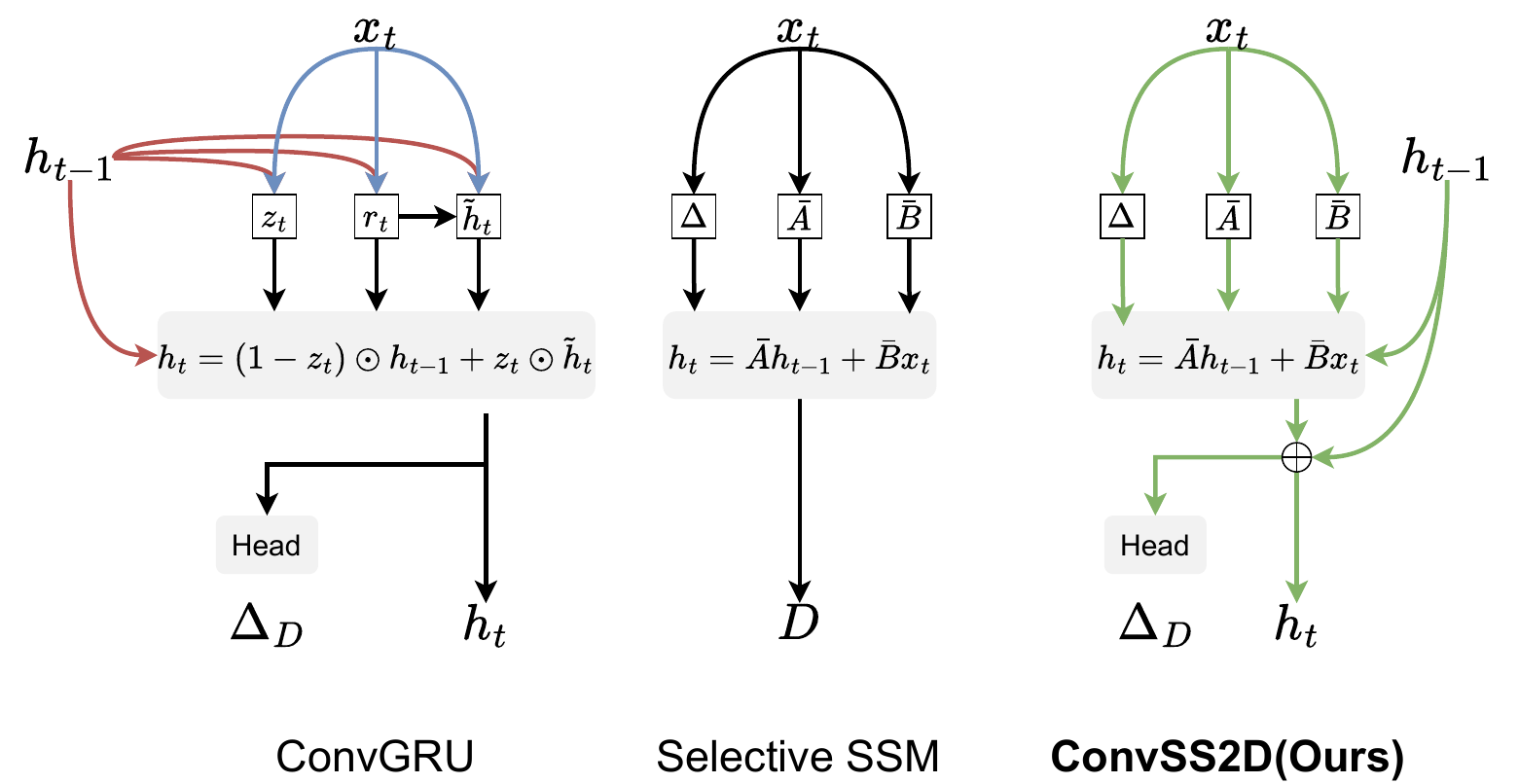}
  \caption{Conceptual comparison. The Gated Recurrent Unit (GRU) relies on multiple non-linear gates and candidate states $\tilde{h}_t$ to update the hidden state ${h}_t$. Its complex gating mechanism introduces non-linear recursion that is difficult to analyze for long sequences. The Selective SSM streamlines this into a linear recurrence. By dynamically generating parameters from the input $x_t$, the Selective SSM maintains "input-dependent selectivity" to adaptively modulate information flow. We leveraged the characteristics of selective SSM to design ConvSS2D, enabling the adaptation iterative process.}
  \label{fig:ss2d_vs_gru}
\end{figure}

Recent advances have sought to bridge monocular vision foundation models (VFMs)~\cite{ren2026anydepth} with stereo geometry for robust underwater adaptation. StereoAdapter~\cite{StereoAdapter} integrates a LoRA-adapted encoder with GRU-based iterative refinement, achieving parameter-efficient domain transfer and demonstrating promising results on underwater benchmarks. However, two key challenges remain for practical underwater deployment: (i) \emph{further improving} the efficiency and accuracy of iterative disparity refinement, particularly in large-disparity and textureless regions prevalent in underwater scenes, and (ii) \emph{bridging the synthetic-to-real gap} given the scarcity of diverse real-world underwater stereo data with accurate ground-truth annotations.

Our motivation is to advance underwater stereo depth estimation along both the architectural and data dimensions while maintaining the parameter-efficient adaptation paradigm. Concretely, we seek to explore alternative update mechanisms that can capture long-range spatial dependencies more effectively, and to construct a large-scale synthetic dataset that better covers the diversity of real underwater conditions including varying optical parameters and camera configurations.

To this end, we propose \textbf{StereoAdapter-2}, a framework that advances underwater stereo depth estimation through architectural innovation and data scaling. \emph{Architecturally}, we introduce the ConvSS2D operator built upon selective state space models~\cite{mamba,vmamba}, which employs a four-directional scanning strategy that naturally aligns with epipolar geometry while capturing vertical structural consistency, enabling efficient long-range spatial propagation at linear computational complexity. \emph{On the data side}, we construct \textbf{UW-StereoDepth-80K}, a large-scale synthetic underwater stereo dataset generated through a two-stage pipeline combining semantic-aware style transfer via Atlantis~\cite{atlantis} and geometry-consistent novel view synthesis via NVS-Solver~\cite{nvssolver}, systematically varying baselines, attenuation coefficients, and scattering parameters to emulate diverse ROV configurations. Combined with dynamic LoRA adaptation inherited from StereoAdapter, our framework achieves state-of-the-art zero-shot performance on underwater benchmarks, with 17\% improvement on TartanAir-UW and 7.2\% on SQUID, while real-world deployment on the BlueROV2 platform validates practical applicability.

The main contributions of this work are summarized as follows:
\begin{itemize}
    \item We introduce the \textbf{ConvSS2D} update operator built upon selective state space models, replacing ConvGRU with a four-directional scanning strategy that captures both horizontal epipolar constraints and vertical structural consistency, enabling efficient long-range spatial propagation within a single refinement step.
    \item We construct \textbf{UW-StereoDepth-80K}, a large-scale synthetic underwater stereo dataset featuring diverse baselines and optical parameters through a two-stage generative pipeline, providing a rigorous foundation for training data-hungry stereo networks.
    \item We achieve state-of-the-art zero-shot performance on underwater benchmarks including TartanAir-UW and SQUID, with real-world validation on the BlueROV2 platform demonstrating robust generalization from synthetic training to real underwater scenes.
\end{itemize}
  
\section{Related Work}
\paragraph{Deep Stereo Matching}
Early deep stereo matching methods mainly relied on CNN-based cost volume aggregation \cite{csm,MC-CNN-acrt,Sgm-nets,spyropoulos2014learning,Ndr}, where stereo correspondence is modeled by constructing and processing cost volumes using 2D or 3D convolutional architectures \cite{CoEx,edgestereo,gcnet,GwcNet,HD3,hsm,iResNet,PSMNet,Ga-net,parameterized,aanet,waveletstereo,cfnet,uasnet,sednet,pcwnet}. However, despite these advances, CNN-based cost aggregation remains fundamentally constrained by explicit cost volume construction, motivating iterative optimization-based stereo methods that bypass explicit aggregation and enable efficient refinement on high-resolution representations \cite{Raft-stereo, Any-stereo, CREStereo, CREStereo++,Eai-stereo, edgestereo, DLNR, IGEV-Stereo, Mc-stereo, Mocha-stereo, Selective-stereo, Orstereo, XR-Stereo, Igev++, ICGNet,StereoAdapter}. The ViT architecture transforms the stereo matching problem into a sequence-to-sequence problem \cite{ELFNet, Chitransformer, croco_v2, CSTR, dynamicstereo, goat, STTR, gmflow}. It uses self-attention and cross-attention \cite{attention} mechanisms with positional encoding to model global context and establish correspondences between stereo views, achieving competitive performance.

\paragraph{Underwater depth estimation and datasets}

Unlike terrestrial scenarios, obtaining accurate and dense ground-truth disparity annotations in underwater environments is extremely difficult, as active sensors such as LiDAR are unreliable underwater and large-scale data collection is costly~\cite{revised_underwater}.  Early underwater datasets, such as FLSea-Stereo \cite{flsea}, lacked accurate Stereo Disparity annotations. UWStereo proposed a high-quality synthetic underwater stereo matching dataset \cite{uwstereo}, but its scene complexity still falls short of real-world underwater scenarios. 

Beyond data limitations, underwater stereo matching itself remains highly challenging. Light scattering, absorption, and refraction significantly reduce photometric consistency between views~\cite{revised_underwater}, making reliable matching difficult. To address these challenges, UWStereo proposed an enhancement module to better perceive geometric structures~\cite{uwstereo}, while UWNet and Fast-UWNet introduced attention mechanisms and 1D–2D cross-search strategies to mitigate underwater image distortions~\cite{UWNet}. However, these approaches rely on carefully designed, domain-specific modules for adaptation, which limits their generalization ability and scalability across diverse underwater conditions.

\paragraph{State Space Model}
State-space model (SSM) has become an efficient alternative to Transformers in sequence modeling \cite{Vision_mamba}. SSM can efficiently model long-range dependencies, and their complexity is linear or near-linear with sequence length. Unlike gated recurrent architectures, SSM relies on structured state evolution, thus achieving stable and scalable sequence processing \cite{long_sequence_modeling, diagonal,combining}. Early SSM-based methods, such as S4 \cite{s4}, improved computational efficiency by parameterizing the state transition matrix and reconstructing sequence modeling into convolution operations. Mamba further improved SSM by introducing selective scanning with input correlation \cite{mamba, mamba2}. Mamba is a hardware-aware algorithm that parallelizes long sequences within a recurrent computation paradigm \cite{mamba}, effectively alleviating the serial bottleneck of traditional RNNs. Mamba demonstrates stronger long sequence modeling capabilities while maintaining high computational efficiency.

Following the successful application of SSM in sequence modeling, recent research has extended them to vision tasks \cite{Mamba-nd, localmamba, vim, vmamba}. Vim adopts a ViT-style architecture and addresses the problem of unidirectionality and lack of positional information in SSM by introducing bidirectional processing and positional embedding \cite{vim, vit}. Vmamba further points out that visual understanding requires modeling that considers spatial structure and global relevance \cite{vmamba}, and proposes the SS2D module, which scans images along multiple spatial directions to capture spatial dependencies. Subsequent research continues to explore improved scanning strategies and scanning direction designs to better utilize the spatial context in visual data \cite{efficientvmamba, Simba, plainmamba, Squeeze-and-excitation}.

The scarcity of underwater scene datasets and the inherent characteristics of SS2D \cite{vmamba} closely match the epipolar geometry in stereo matching tasks, inspiring our approach. We leverage the rich representations learned in the pre-trained model while utilizing LoRa for efficient parameter fine-tuning and domain adaptation. Replacing the traditional GRU-based update module with an SSM-based module enables effective underwater depth estimation.

\begin{figure*}[t]
  \centering
  \includegraphics[width=\linewidth]{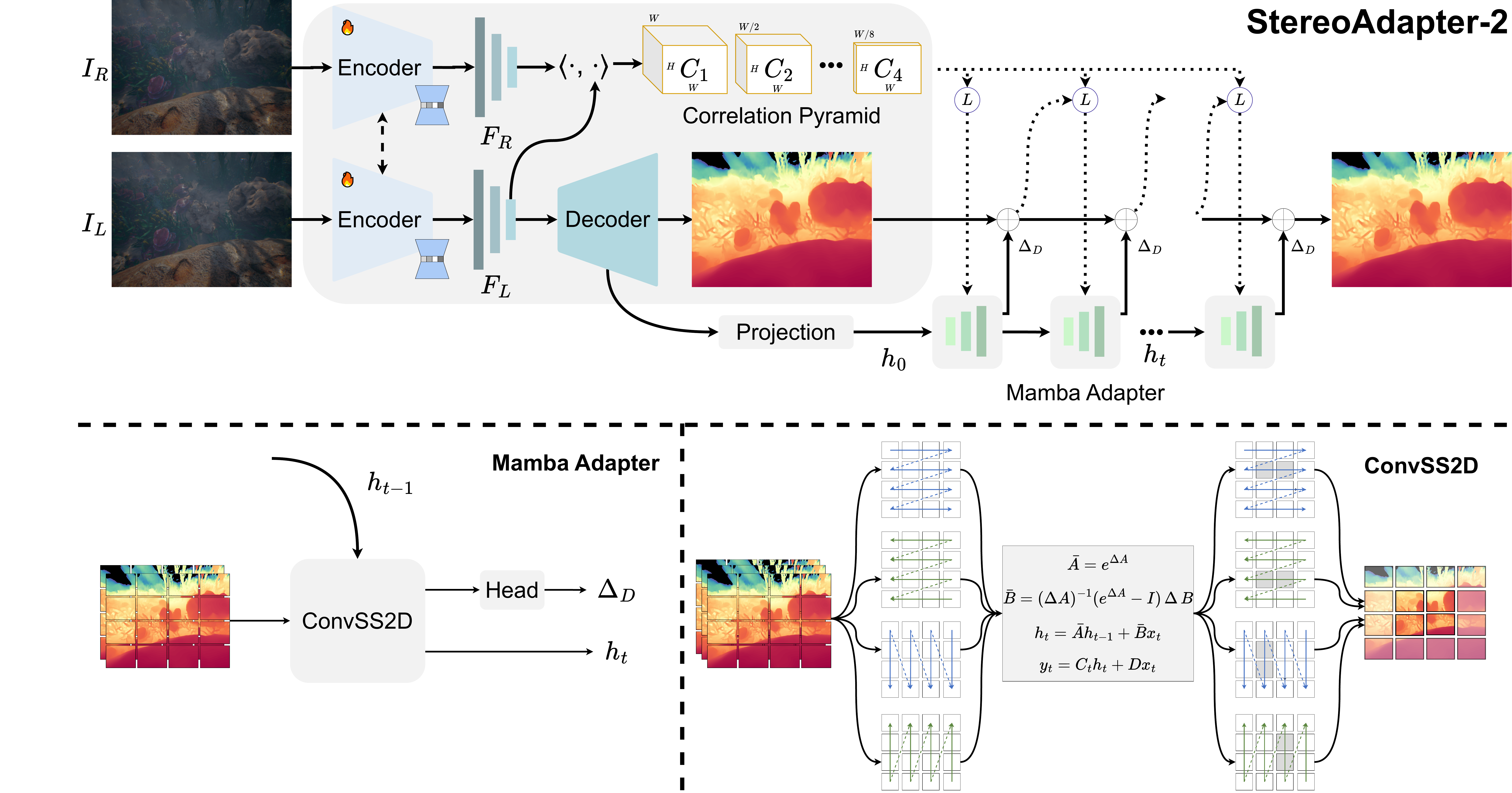}
  \caption{Detailed architecture of the StereoAdapter-2: Our model iteratively refines disparity by integrating a Mamba Adapter. The refinement step is powered by the ConvSS2D operator, which enables adaptive and long-range spatial information propagation through multi-directional selective scanning.}
  \label{fig:overview}
\end{figure*}

\section{Preliminaries}
The SSM is a type of continuous-time latent state model, which defines a mapping from a one-dimensional function or sequence $u(t) \in \mathbb{R}$ to an output $y(t) \in \mathbb{R}$ through an implicit latent state $h(t) \in \mathbb{R}^N$, as given in Eq.~\ref{ssm}.

\begin{equation}
\begin{aligned}
\mathbf{h}'(t) &= \mathbf{A} \mathbf{h}(t) + \mathbf{B} u(t), \\
y(t) &= \mathbf{C} \mathbf{h}(t) + D u(t),
\label{ssm}
\end{aligned}
\end{equation}
where $\mathbf{A}, \mathbf{B}, \mathbf{C}, D$ are the learned parameters, and for the sake of explanation, we omit parameter $D$.

For SSM model training, we discretize the parameters of the continuous-time system. As shown in Eq.~\ref{discretize}, the continuous-time parameters $\mathbf{A}$ and $\mathbf{B}$ are discretized using a zero-order hold (ZOH), where $\Delta$ denotes the discretization time step.

\begin{equation}
\begin{aligned}
\bar{\mathbf{A}} &= e^{\Delta \mathbf{A}}, \\
\bar{\mathbf{B}} &= (\Delta \mathbf{A})^{-1} \left( e^{\Delta \mathbf{A}} - \mathbf{I} \right)\, \Delta\, \mathbf{B}.
\label{discretize}
\end{aligned}
\end{equation}

After discretization, the entire model can be computed using linear recurrence and global convolution. Global convolution computation can be efficiently parallelized, and efficient autoregressive inference can be performed through linear recurrence.

\begin{equation}
\begin{alignedat}{1}
\bar{\mathbf{K}} &=
\left(
\mathbf{C}\bar{\mathbf{B}},
\;\mathbf{C}\bar{\mathbf{A}}\bar{\mathbf{B}},
\;\ldots,
\;\mathbf{C}\bar{\mathbf{A}}^{L-1}\bar{\mathbf{B}}
\right), \\
\mathbf{y} &= \mathbf{x} * \bar{\mathbf{K}},
\end{alignedat}
\end{equation}
where $L$ is the length of the input sequence, and $\bar{\mathbf{K}} \in \mathbb{R}^{L}$ denotes the structured convolutional kernel. This formulation provides a general view of state space models, which we later reinterpret as structured spatial state recursion for iterative disparity refinement.

\section{The Proposed Method}

\subsection{Overview}
We propose StereoAdapter-2, which uses a monocular depth foundation model to guide stereo disparity estimation, as shown in Fig. \ref{fig:overview}. Our framework adopts a unified architecture that integrates Depth Anything 3 \cite{da3} as both the feature encoder and monocular depth estimator. To efficiently adapt the pretrained Depth Anything 3 to stereo matching in underwater scenes, we employ LoRA \cite{StereoAdapter}, which enables efficient parameter fine-tuning while maintaining the rich representations learned from large-scale pre-training. Monocular depth estimation is utilized for disparity initialization to accelerate convergence. To iterate disparity estimation, we replace the traditional GRU-based update module with a Selective SSM module and enhance the learned gating mechanism. This design leverages the long-range spatial modeling capabilities of the SSM while retaining the adaptive memory control of the cyclic unit.

\subsection{Feature Extraction}

We first extract features $F_L$ and $F_R$ using the powerful depth foundation model Depth Anything 3 \cite{da3}. We extract multi-scale representations from four intermediate Transformer layers ${T^1, T^2, T^3, T^4}$ to capture details and semantic information at different levels. Meanwhile, for underwater scene domain adaptation, we fine-tune the encoder following the approach of StereoAdapter \cite{StereoAdapter}.

\subsection{Correlation Pyramids Building}
We constructed a correlation pyramid to encode the visual similarity between pairs of stereo images. Unlike the optical flow of a 4D correlator that needs to cover all pixel pairs, stereo matching using calibrated images restricts the correspondences to the horizontal direction.

Given the features $f_l^1, f_r^1 \in \mathbb{R}^{H \times W \times D}$ extracted from $F_L$ and $F_R$, we compute the correlation volume by calculating the inner product between features with the same $y$ coordinate, following Eq~\eqref{correlation}.

\begin{equation}
C_{i j k} = \sum_{d} f_{l,i j d}^{1} \cdot f_{r,i k d}^{1},
\quad C \in \mathbb{R}^{H \times W \times W},
\label{correlation}
\end{equation}
where $i$ represents the row index of the left image, $j$ represents the column index of the left image, and $k$ represents the column index of the right image. To capture the correspondence between fine-grained and large displacements, we construct a four-layer correlation pyramid ${C^{(l)}}_{l=1}^{4}$ by repeatedly applying average pooling with a kernel size of 2 along the last dimension, where the $l$-th layer has a dimension of $H \times W \times W/2^{l-1}$, providing a progressively larger receptive field while maintaining spatial resolution. In each refinement iteration, given the current disparity estimate $d$, we perform a lookup operation using linear interpolation to retrieve the correlation values at integer offsets ${d-r, \ldots, d+r}$ from each pyramid layer, and concatenate the retrieved values from all layers to form the correlation features input to the update operator.
  
\subsection{Iterative Disparity Estimation}
  
Following RAFT-Stereo \cite{Raft-stereo}, we adopt an iterative refinement framework to progressively estimate disparity. Given an initial disparity estimate $D_0$, we iteratively update it through $L$ iterations: $D_0, D_1, ..., D_L$. However, instead of using ConvGRU,  We propose ConvSS2D as the core operator. Firstly, the long-range dependency modeling of ConvSS2D is achieved through sequential state recursion \cite{mamba,mamba2,vmamba}, without requiring multiple layers of convolutions to expand the receptive field. Specifically, the state update at spatial location $t$ follows Eq~\eqref{eq:ssm_update}.
\begin{equation}
h_t = \bar{A} h_{t-1} + \bar{B} x_t,
\label{eq:ssm_update}
\end{equation}
where $h_t$ denotes the hidden state at spatial position $t$ along a given scan direction, $h_{t-1}$ represents the propagated state from the previous position, and $x_t$ is the input feature at the current location. The discretized state transition matrix $\bar{A}$ governs how information propagates sequentially across spatial positions, while $\bar{B}$ controls how the input features are incorporated into the state update. As a result, information can be propagated over long spatial extents through directional scans, allowing features at distant locations to influence each other within a single refinement step. Owing to the inherent long-range propagation capability of ConvSS2D, we discard the traditional context encoder and directly project decoder features to initialize the hidden state $h_0$

\paragraph{Input-dependent Selectivity}
A key limitation of ConvGRU lies in its inductive bias for spatial information propagation. Although its gating functions are conditioned on the input, the update is implemented through local convolutional kernels, resulting in predominantly local and isotropic information aggregation within each refinement step. In contrast, ConvSS2D introduces input-dependent selectivity through dynamically computed parameters
$\Delta$, $B$, and $C$. These parameters are generated from the input
features $\mathbf{x}$ via linear projections following Eq~\eqref{eq:Selectivity}.
\begin{equation}
  \Delta = \mathrm{softplus}(W_\Delta \mathbf{x_t}), \quad
  B = W_B \mathbf{x_t}, \quad
  C = W_C \mathbf{x_t},
  \label{eq:Selectivity}
\end{equation}
where $W_\Delta$, $W_B$, $W_C$ are learnable projection matrices. This
mechanism enables the model to adaptively modulate: (1) the state update dynamics via $\Delta$, controlling the rate of state evolution; (2) input gating via $B$, selectively incorporating relevant features; and (3) output projection via $C$, emphasizing task-relevant information. Such content-aware processing allows the network to dynamically adjust its behavior based on local image characteristics, such as texture, edges, and occlusion boundaries.

\paragraph{Scanning Strategy}
We extend one-dimensional selective scanning to two dimensions using a four-directional scanning strategy that handles features along both horizontal and vertical directions. This design is particularly suitable for stereo matching, because reliable matching still benefits from aggregating two-dimensional spatial context. The horizontal scan is directly aligned with the epipolar constraint, enabling efficient propagation of disparity information along the scan line. Simultaneously, the vertical scan contributes to consistency across the scan line, captures vertical structure, and normalizes disparity estimation in textureless regions. The outputs from all four scan directions are aggregated to form a comprehensive feature representation that satisfies the inherent geometric constraints of stereo vision.

\subsection{Data Synthesis: UW-StereoDepth-80K}

\begin{figure}[t]
  \centering
  \includegraphics[width=\columnwidth]{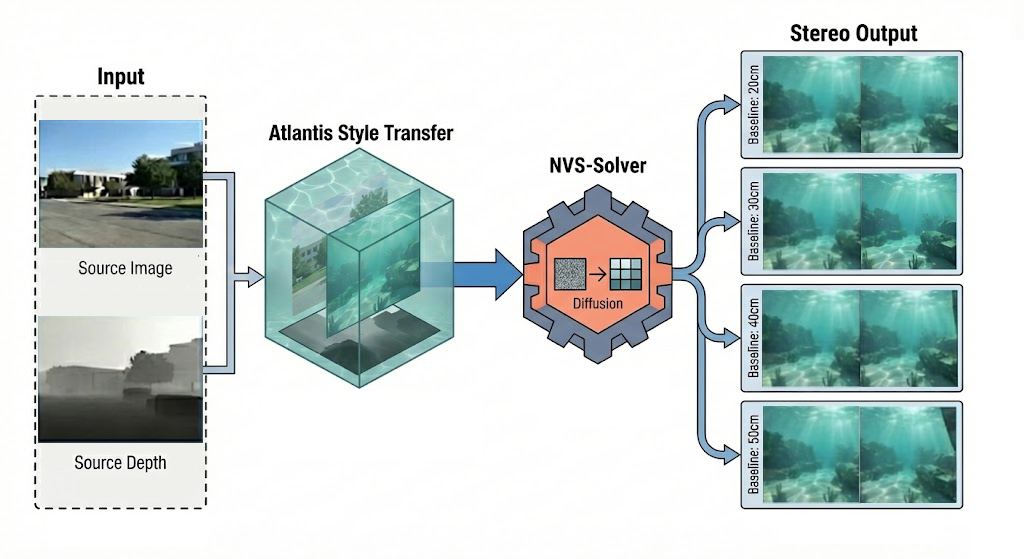}
  \caption{Data synthesis pipeline. Semantic-aware style transfer and geometry-consistent novel view synthesis rendering pipeline for UW-StereoDepth-80K dataset.}
  \label{fig:pipeline}
\end{figure}

To overcome the scarcity of diverse real-world underwater stereo data, we propose a novel two-stage generative data synthesis pipeline. Our approach leverages diffusion models to synthesize high-fidelity underwater stereo pairs from terrestrial RGB-D data. As illustrated in Fig. \ref{fig:pipeline}, our pipeline sequentially applies semantic-aware style transfer and geometry-consistent novel view synthesis.

\paragraph{Underwater Style Transfer}
We utilize Atlantis \cite{atlantis}, a specialized framework for enabling underwater data synthesis via Stable Diffusion, to bridge the photometric domain gap.  Given a terrestrial source image $I_{src}$ and its corresponding source depth map $D_{src}$, Atlantis acts as a style transfer module that hallucinates realistic underwater optical effects, such as wavelength-dependent attenuation, scattering, and turbidity, while preserving the semantic content and geometric structure of the original scene. By conditioning the diffusion process on the source depth $D_{src}$, we ensure that the synthesized underwater imagery maintains structural fidelity to the input, effectively transforming a terrestrial dataset into a diverse underwater domain without losing ground truth geometric labels.

\paragraph{Multi-Baseline Stereo Generation}
To generate stereo correspondences from the stylized monocular images, we employ NVS-Solver \cite{nvssolver}, a video diffusion model designed for zero-shot novel view synthesis. Standard diffusion-based image generation often lacks multi-view geometric consistency. NVS-Solver addresses this by treating the stereo generation task as a view synthesis problem governed by explicit camera extrinsics. Taking the output from the Atlantis stage as the reference view, we synthesize the target right view by conditioning the solver on specific baseline displacements. As shown in the right panel of Fig. \ref{fig:pipeline}, we systematically generate stereo pairs across four distinct baselines: 20cm, 30cm, 40cm, and 50cm. This multi-baseline strategy simulates the diverse camera configurations found in real-world underwater robots, thereby enhancing the model's robustness to scale variations and disparity ranges during training.

\paragraph{Dataset Construction}
By cascading Atlantis and NVS-Solver, we convert large-scale terrestrial RGB-D datasets into a synthetic underwater stereo benchmark. Each stereo pair in our generated subset is synthesized at a resolution of $640 \times 480$. The resulting dataset features physically plausible underwater appearance, consistent stereo geometry, and dense ground truth disparity, providing a rigorous foundation for training data-hungry stereo matching networks. UW-StereoDepth-80K is constructed by merging our newly generated diffusion-based samples with the existing UW-StereoDepth-40K dataset \cite{StereoAdapter}. The final consolidated dataset comprises 80,000 high-quality stereo image pairs.

\section{Experiments}

\subsection{Datasets and Metrics}

\paragraph{Training Datasets}

To cover various underwater scenarios, we used our training data based on the \emph{UW-StereoDepth-80K} dataset, which contains about 80K samples generated using NVS-solver~\cite{nvssolver} to synthesize virtual underwater data. For evaluation, we conduct experiments on two underwater datasets. The first is TartanAir-UW, a subset from TartanAir \cite{tartanair} that only consists of 13,583 underwater stereo image pairs. The second is the SQUID dataset \cite{berman2020underwater}, which contains images from four distinct scenes.

\paragraph{Evaluation Dataset and Metrics}
We report standard depth estimation metrics, including Absolute Mean Relative Error (AbsRel), Squared Mean Relative Error (SqRel), Root Mean Square Error (RMSE), and logarithmic RMSE (Log RMSE). In addition, we report accuracy under threshold metrics $\delta_1$, $\delta_2$, and $\delta_3$. The accuracy threshold $\delta_k$ measures the percentage of pixels for which $\max\left(\frac{\hat d_i}{d_i}, \frac{d_i}{\hat d_i}\right) < 1.25^k,$ where $d_i$ and $\hat d_i$ denote the ground-truth and predicted depth values, respectively, and $k \in \{1, 2, 3\}$.

\subsection{Implementation Details}
We trained StereoAdapter-2 on an H100 NVL and deployed it on ROV. Input image resolution is $480 \times 640$ and normalized to \([0,1]\). We initialize the feature encoder with Depth Anything 3 (ViT-B) \cite{da2} pretrained weights. We perform 22 iterations during training and 32 during inference. For LoRA settings, we follow the StereoAdapter \cite{StereoAdapter} settings, LoRA rank $r=16$, sparsity threshold $\kappa_{\max}=0.005$, and regularization weight $\lambda=1 \times 10^{-4}$. The sparse phase activates at 50\% of training. Our method uses the loss function $\mathcal{L}_{\text{disparity}}$ and $\mathcal{L}_{\text{sparse}}$, and the weight ratio is set to $1:1$. The model is trained using the AdamW optimizer with a learning rate of $1 \times 10^{-4}$ and weight decay of $1 \times 10^{-5}$. We employ the OneCycleLR scheduler for 100K iterations. Regarding data augmentation, we used strategies consistent with RAFT-Stereo \cite{Raft-stereo}, including saturation enhancement and random scaling.

\subsection{Main Results}

\begin{figure*}[t]
  \centering
  \includegraphics[width=\textwidth]{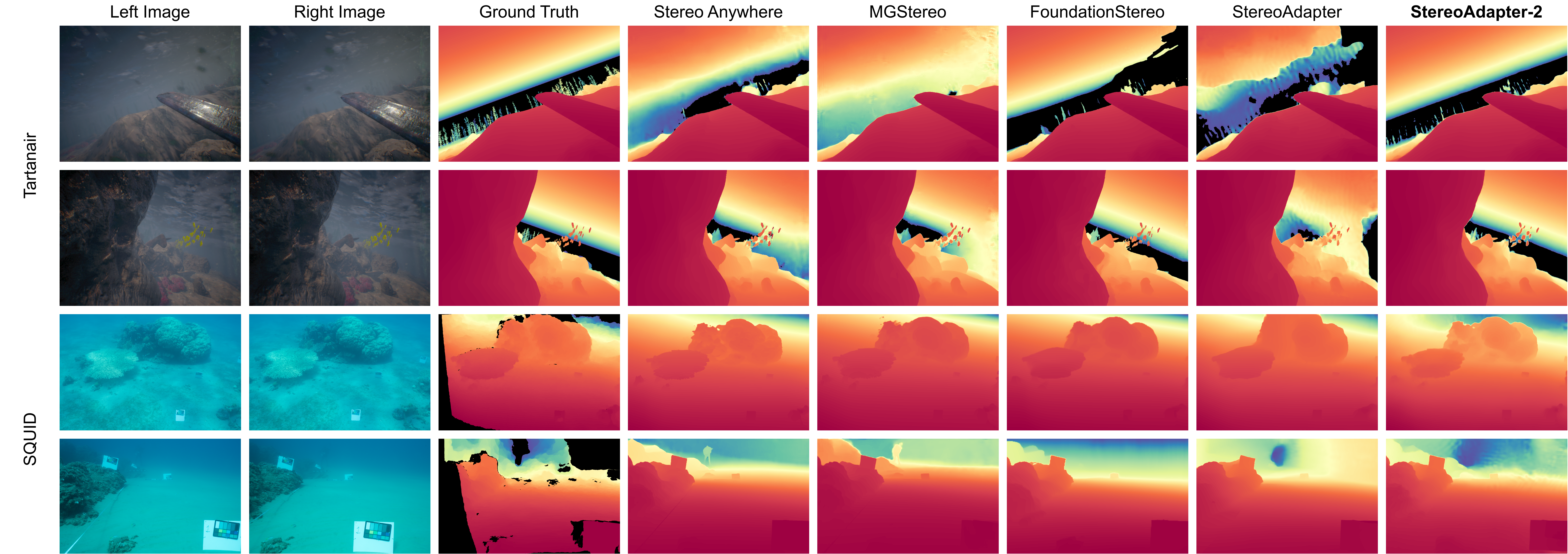}
  \caption{Qualitative results of zero-shot stereo depth estimation}
  \label{fig:visualization}
\end{figure*}

\input{tables/quant_tartanair_occ}

Our experiments demonstrate that the proposed StereoAdapter-2, trained on the UW-StereoDepth-80K dataset, achieves state-of-the-art zero-shot performance across both TartanAir Underwater and SQUID benchmarks. As summarized in Tables~\ref{tab:quant_tartanair_occ} and~\ref{tab:quant_squid}, our approach consistently outperforms existing stereo matching methods without any fine-tuning on the target domains.

As shown in Table~\ref{tab:quant_tartanair_occ}, StereoAdapter-2 achieves superior zero-shot performance on the TartanAir Underwater subset, obtaining the lowest REL (0.0440) and RMSE (2.4038), along with the highest A1 accuracy (96.76\%). Compared to our prior StereoAdapter trained on UW-StereoDepth-40K, StereoAdapter-2 reduces REL by 16.5\% and RMSE by 17.0\%, demonstrating both the effectiveness of our adapter architecture and the benefits of scaling the training dataset.

Table~\ref{tab:quant_squid} presents zero-shot evaluation on the real-world SQUID dataset. StereoAdapter-2 attains the best overall performance with an RMSE of 1.7481 and the lowest REL of 0.0705, reducing RMSE by 7.2\% compared to the previous StereoAdapter while achieving leading accuracy across all $\delta$ thresholds (A1: 94.25\%, A2: 97.65\%, A3: 98.62\%). These results highlight the strong zero-shot generalization capability of StereoAdapter-2 from synthetic training data to real-world underwater scenes.

As shown in Figure~\ref{fig:visualization}, StereoAdapter-2 generates substantially more accurate and visually coherent depth maps than baseline methods, with better scale estimation for far range details.

In summary, these findings validate that our StereoAdapter-2 architecture, combined with the UW-StereoDepth-80K dataset, enables robust zero-shot stereo depth estimation in diverse underwater environments.

\begin{figure*}[t]
  \centering
  \includegraphics[width=\textwidth]{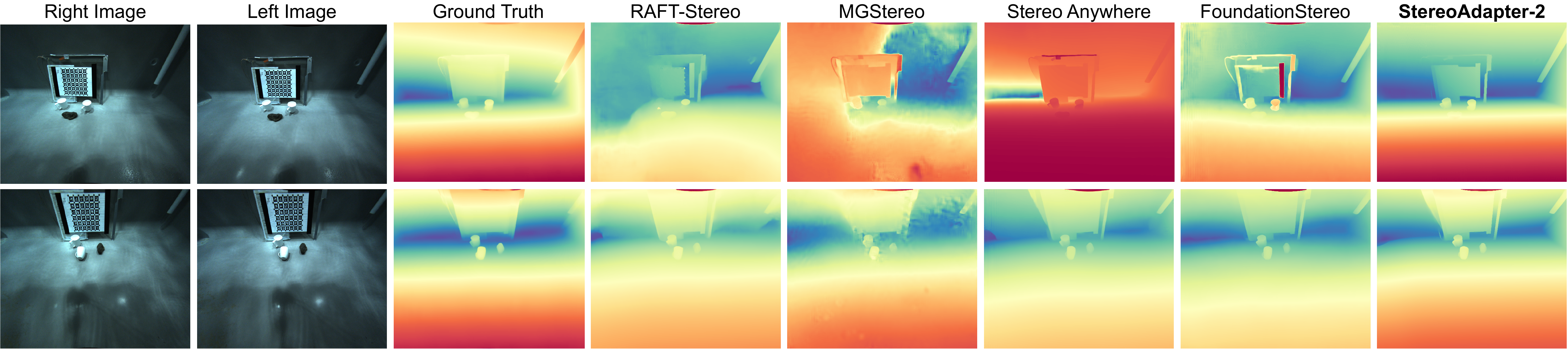}
  \caption{Qualitative results of zero-shot underwater stereo depth estimation were obtained by deploying the model on a robotic platform.}
  \label{fig:real_world_visualization}
\end{figure*}

\input{tables/quant_squid}

\input{tables/quant_bluerov2}

\subsection{Real-World Evaluation}
\begin{figure}[h]
  \centering
  \includegraphics[width=\columnwidth]{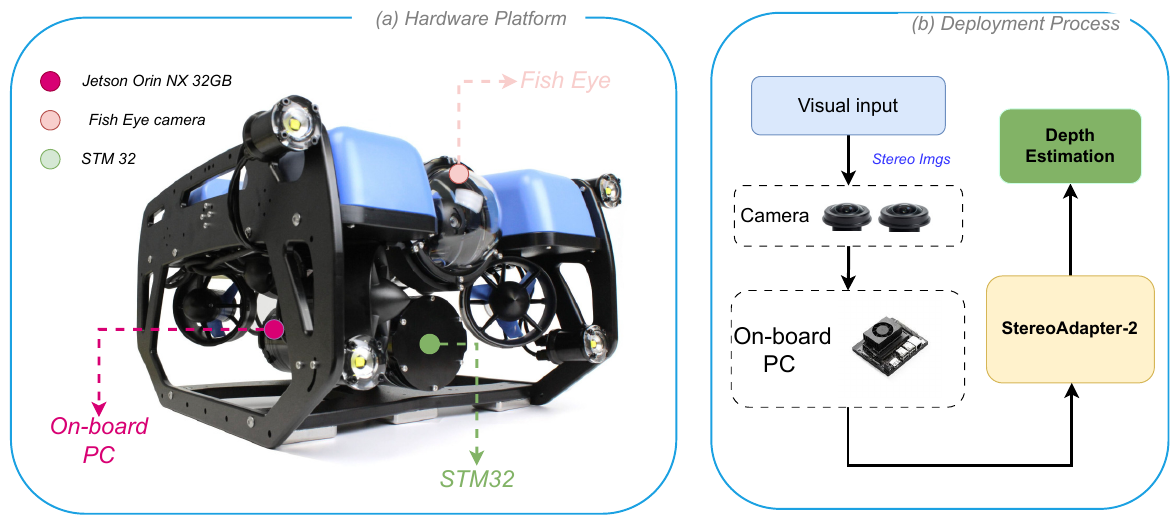}
  \caption{Hardware platform for real world experiments.}
  \label{fig:realworld}
\vspace{-2em}
\end{figure}

\paragraph{Hardware Configuration}

As shown in Figure \ref{fig:realworld}, We validate our approach using a BlueROV2 platform equipped with an NVIDIA Jetson Orin NX (32GB) for onboard computation. Low-level motion control is delegated to an STM32 microcontroller. Visual input is captured by a pair of fisheye cameras mounted in a stereo arrangement; we apply offline rectification to transform the raw fisheye frames into a standard pinhole geometry prior to network inference.

\paragraph{Scene Setup and Data Collection}
We conduct all trials in a controlled indoor water tank environment. To emulate realistic underwater navigation scenarios, we arrange glass containers and irregularly shaped stones into 5 distinct spatial layouts representing different levels of clutter complexity. The robot is then teleoperated through 3 separate navigation routes per layout, yielding a total of 15 time-aligned binocular recordings. All visual data and timestamps are logged directly on the Jetson platform.

\paragraph{Ground-Truth Acquisition}
Before experimentation, we construct a geometrically calibrated 3D reference model of the tank interior. During each trial, camera poses are recovered by detecting AprilTags (family 16h5) and solving the corresponding pose estimation problem. These poses are subsequently aligned with the pre-scanned model, enabling us to project the geometry onto the left view and obtain per-pixel depth references. Regions without valid surface intersections are excluded from subsequent analysis.
\paragraph{Evaluation Protocol}
Every method under comparison receives the same rectified image pairs at a uniform resolution with consistent pre-processing. When a model produces disparity output, we recover absolute depth via the known stereo geometry parameters. Performance is quantified using established depth metrics: Absolute Relative Error (REL), Squared Relative Error (SQ REL), Root Mean Squared Error (RMSE), Logarithmic RMSE, and the threshold accuracy A1. All statistics are computed exclusively over valid pixels and aggregated across the full set of recordings.

\paragraph{Results}
The proposed method achieves better performance, as shown in table \ref{tab:quant_bluerov2}, reaching a REL of 0.1023, an RMSE of 1.7164, and A1 accuracy of 92.56\%. Relative to other baselines, our proposed model exhibits consistent gains in both precision and stability across diverse underwater obstacle arrangements.

\subsection{Ablation Study}
\input{tables/ablation_model}
\input{tables/ablation_hyperparameters}
\input{tables/ablation_ss2d}
\input{tables/ablation_scan}
\input{tables/inference_time}

Table \ref{tab:ablation_model} shows ablation experiments for different components of our model, including the use of a pre-trained model, monocular disparity initialization, ConvSS2D, and the context encoder. Table \ref{tab:ablation_training} shows the ablation experiments performed on different hyperparameter settings during model training. 

Table \ref{tab:ablation_ssm} further analyzes the impact of key SSM hyperparameters in ConvSS2D, including the state dimension $d_{\text{state}}$ and the SSM expansion ratio. We observe that increasing $d_{\text{state}}$ progressively improves model performance, with $d_{\text{state}}=16$ achieving the best REL and RMSE scores. However, this comes at the cost of increased computational overhead and reduced throughput. In contrast, increasing the SSM expansion ratio beyond $1.0$ leads to significant performance degradation. Considering the trade-off between accuracy and efficiency, we choose $d_{\text{state}}=4$ with an SSM ratio of $1.0$ as the default configuration, which achieves competitive performance while maintaining high throughput.
Table \ref{tab:ablation_scanning} investigates the impact of different SS2D scanning modes. Compared to unidirectional and bidirectional scanning, the cross-scanning strategy consistently achieves better model performance while maintaining the number of similar parameters and FLOPs. This indicates that reliable matching still benefits from the aggregation of two-dimensional spatial context.

\section{Test-Time Efficiency}
\label{sec:test_time}
We evaluate on an on-board {Jetson Orin NX 32GB} in {MaxN} mode with TensorRT, batch size~1, and input resolution {$640{\times}320$}. With identical pre/post-processing for predictions. We report per-frame end-to-end latency in milliseconds (ms).

As shown in table \ref{tab:orin_nx_latency} FoundationStereo and Stereo Anywhere both adopt DepthAnythingV2-L as their encoder backbone, with FoundationStereo incurring additional overhead from its transformer-based feature refinement module. MGStereo, while using a lighter encoder, involves multi-stage disparity fusion and iterative refinement, which contributes to its latency. In contrast, StereoAdapter-2 achieves the lowest latency of 1102\,ms by employing a LoRA-adapted DepthAnythingV3-B encoder and replacing conventional recurrent updates with ConvSS2D, which accelerates the disparity refinement process while maintaining accuracy.

\section{Limitations and Future Work}
\label{sec:limitations}
Despite these advances, limitations remain. The synthetic-to-real domain gap persists under extreme underwater conditions, such as severe turbidity, strong backscatter, or rapidly varying illumination, where the diversity of our training data may not fully capture real-world complexity. Furthermore, while our method achieves strong per-frame accuracy, temporal consistency in continuous deployment remains challenging—consecutive depth predictions may exhibit flickering or instability. Future work will focus on incorporating temporal modeling to ensure prediction stability across consecutive frames, as well as exploring tighter integration with downstream robotic tasks, such as grasp point prediction for underwater manipulation.

\section{Conclusion}
\label{sec:conclusion}
We present StereoAdapter-2, a novel framework for underwater stereo depth estimation by introducing the ConvSS2D operator, built upon selective state space models, our method enables efficient long-range spatial propagation through a four-directional scanning strategy. To address the scarcity of diverse underwater stereo data, we constructed UW-StereoDepth-80K through a two-stage generative pipeline, combining semantic-aware style transfer and geometry-consistent novel view synthesis, enabling systematic variation for underwater images. Combined with dynamic LoRA adaptation, our framework achieves state-of-the-art zero-shot performance on underwater benchmarks, with \textbf{17\%} improvement on TartanAir-UW and \textbf{7.2\%} on SQUID compared to prior methods. Real-world deployment on the BlueROV2 platform further validates the practical applicability of our approach.

\clearpage

\bibliographystyle{plainnat}
\bibliography{references}

\clearpage

\input{appendix}

\end{document}

%% file: tables/quant_tartanair_occ.tex
\begin{table*}[t]
\centering
\caption{Quantitative comparison of zero-shot stereo depth estimation on the TartanAir underwater subset.
All methods are evaluated under the same protocol using standard depth metrics.}

\label{tab:quant_tartanair_occ}
\resizebox{\textwidth}{!}{%
\begin{tabular}{l l c c c c c c c}
\toprule
\textbf{Method} 
& \textbf{Training Set} 
& \textbf{Rel$\downarrow$}
& \textbf{SqRel$\downarrow$}
& \textbf{RMSE$\downarrow$}
& \textbf{Log RMSE$\downarrow$}
& \textbf{A1$\uparrow$}
& \textbf{A2$\uparrow$}
& \textbf{A3$\uparrow$} \\
\midrule
LEAStereo \cite{LEAStereo}        & Scene Flow               & 0.1099 & 1.3898 & 4.5610 & 0.2063 & 0.8929 & 0.9512 & 0.9761 \\
PSMNet \cite{PSMNet}             & Scene Flow               & 0.0884 & 0.8699 & 3.9721 & 0.1804 & 0.9122 & 0.9627 & 0.9804 \\
AANet \cite{aanet}               & Scene Flow               & 0.6096 & 8.3687 & 13.0542 & 0.9903 & 0.2598 & 0.3451 & 0.3888 \\
GwcNet \cite{GwcNet}             & Scene Flow               & 0.1013 & 1.2965 & 4.1829 & 0.1855 & 0.9085 & 0.9612 & 0.9801 \\
ACVNet \cite{ACVNet}             & Scene Flow               & 0.0970 & 1.1335 & 3.9985 & 0.1803 & 0.9063 & 0.9612 & 0.9813 \\
RAFT-Stereo \cite{Raft-stereo}   & Scene Flow               & 0.0814 & 0.7342 & 4.0423 & 0.1703 & 0.9030 & 0.9612 & 0.9832 \\
HSMNet \cite{hsm}                & Scene Flow               & 0.9856 & 12.3768 & 15.2865 & 4.5961 & 0.0000 & 0.0000 & 0.0000 \\
TiO-Depth \cite{tiodepth}        & KITTI2012                & 0.7194 & 8.6479 & 13.4635 & 1.6967 & 0.0053 & 0.0096 & 0.0550 \\
FoundationStereo \cite{FoundationStereo} 
                                  & FoundationStereo dataset & 0.0542 & 0.6701 & 2.9644 & 0.1358 & 0.9302 & 0.9701 & 0.9779 \\
Stereo Anywhere \cite{Stereoanywhere}
                                  & Scene Flow               & 0.0592 & 0.5098 & 3.1572 & 0.1544 & 0.9442 & \textbf{0.9787} & 0.9889 \\
CREStereo \cite{CREStereo}        & ETH3D                    & 2.5746 & 9.8789 & 8.4526 & 5.1297 & 0.4890 & 0.5732 & 0.7001 \\
\midrule
\rowcolor[gray]{0.95}
StereoAdapter \cite{StereoAdapter}
                                  & UW-StereoDepth-40K       & 0.0527 & 0.5167 & 2.8947 & 0.1371 & 0.9467 & 0.9701 & 0.9753 \\
\rowcolor[rgb]{ .886,  .937,  .855}
\textbf{StereoAdapter-2 (Ours)}   & UW-StereoDepth-80K       & \textbf{0.0440} & \textbf{0.4312} & \textbf{2.4038} & \textbf{0.1198} & \textbf{0.9676} & 0.9704 & \textbf{0.9890} \\
\bottomrule
\end{tabular}}
\end{table*}

%% file: tables/quant_squid.tex
\begin{table*}[t]
\centering
\caption{
Zero-shot evaluation on SQUID dataset. 5 Datasets$^\ast$ refers to Scene Flow~\cite{mayer2016large},
Sintel~\cite{butler2012naturalistic},
ETH3D~\cite{schoeps2017cvpr},
InStereo2K~\cite{Bao2020InStereo2KAL},
and CREStereo~\cite{li2022practical}.
}

\label{tab:quant_squid}
\resizebox{\textwidth}{!}{%
\begin{tabular}{l l c c c c c c c}
\toprule
\textbf{Method} 
& \textbf{Training Set} 
& \textbf{Rel$\downarrow$}
& \textbf{SqRel$\downarrow$}
& \textbf{RMSE$\downarrow$}
& \textbf{Log RMSE$\downarrow$}
& \textbf{A1$\uparrow$}
& \textbf{A2$\uparrow$}
& \textbf{A3$\uparrow$} \\
\midrule
LEAStereo \cite{LEAStereo}        & Scene Flow                    & 0.5574 & 3.9434   & 5.4659  & 0.4335 & 0.6512 & 0.8042 & 0.8869 \\
PSMNet \cite{PSMNet}           & Scene Flow                    & 0.5182 & 7.1404   & 4.9186  & 0.5902 & 0.7139 & 0.7999 & 0.8311 \\
AANet \cite{aanet}            & Scene Flow                    & 7.4801 & 314.1577 & 34.7612 & 1.8994 & 0.0602 & 0.1087 & 0.1570 \\
GwcNet \cite{GwcNet}           & Scene Flow                    & 0.2294 & 1.2275   & 3.0003  & 0.3799 & 0.7423 & 0.8517 & 0.9005 \\
ACVNet \cite{ACVNet}           & Scene Flow                    & 1.6030 & 65.6518  & 10.3828 & 0.7293 & 0.7019 & 0.7925 & 0.8321 \\
RAFT-Stereo \cite{Raft-stereo}     & Scene Flow                    & 0.0831 & 0.6946   & 1.9625  & 0.1441 & 0.9235 & 0.9634 & 0.9835 \\
HSMNet \cite{hsm}           & Scene Flow                    & 0.9772 & 7.2766   & 8.2301  & 4.0887 & 0.0000 & 0.0000 & 0.0000 \\
CREStereo \cite{CREStereo}        & ETH3D                         & 2.5746 & 9.8789   & 8.4526  & 5.1297 & 0.4890 & 0.5732 & 0.7001 \\
IGEV-Stereo \cite{IGEV-Stereo}      & 5 Datasets$^\ast$ + TartanAir & 0.0932 & 1.4685   & 2.4741  & 0.1523 & 0.9346 & 0.9712 & 0.9820 \\
Selective IGEV \cite{Selective-stereo}   & 5 Datasets$^\ast$ + TartanAir & 0.0960 & 0.9617   & 1.9268  & 0.1665 & 0.9171 & 0.9555 & 0.9720 \\
GMStereo \cite{gmstereo}        & 5 Datasets$^\ast$ + TartanAir & 3.3442 & 140.3211 & 18.7829 & 1.0219 & 0.5300 & 0.6076 & 0.6578 \\
TiO-Depth \cite{tiodepth}        & KITTI2012                     & 1.3154 & 11.6828  & 7.0930  & 0.8121 & 0.1753 & 0.3346 & 0.5133 \\
FoundationStereo \cite{FoundationStereo} & FoundationStereo dataset      & 0.1095 & 0.7012   & 2.2510  & 0.1584 & 0.8995 & 0.9433 & 0.9501 \\
Stereo Anywhere \cite{Stereoanywhere}  & Scene Flow                    & 0.0952 & 1.1017   & 2.4317  & 0.1586 & 0.9179 & 0.9605 & 0.9763 \\
\midrule
\rowcolor[gray]{0.95}
StereoAdapter \cite{StereoAdapter} 
                      & UW-StereoDepth-40K             & 0.0806 & 0.7082   & 1.8843  & 0.1469 & 0.9413 & 0.9748 & 0.9852 \\
\rowcolor[rgb]{ .886,  .937,  .855}
\textbf{StereoAdapter-2 (Ours)}   & UW-StereoDepth-80K & \textbf{0.0705} & \textbf{0.6396} & \textbf{1.7481} & \textbf{0.1285} & \textbf{0.9425} & \textbf{0.9765} & \textbf{0.9862} \\
\bottomrule
\end{tabular}}
\vspace{-1em}
\end{table*}

%% file: tables/quant_bluerov2.tex
\begin{table}[t]
\centering
\caption{Real-world evaluation on BlueROV2.}
\label{tab:quant_bluerov2}
\resizebox{\linewidth}{!}{%
\begin{tabular}{l c c c c c}
\toprule
\textbf{Method} 
& \textbf{REL$\downarrow$}
& \textbf{SqRel$\downarrow$}
& \textbf{RMSE$\downarrow$}
& \textbf{Log RMSE$\downarrow$}
& \textbf{A1$\uparrow$} \\
\midrule
Stereo Anywhere \cite{Stereoanywhere}     & 0.1218 & 1.0623 & 2.4682 & 0.1673 & 0.8541 \\
FoundationStereo \cite{FoundationStereo}   & 0.1304 & \underline{0.6187} & 2.0893 & 0.1635 & \underline{0.8812} \\
\rowcolor[gray]{0.95}
StereoAdapter \cite{StereoAdapter}
                        & 0.1163 & 0.6794 & 1.9285 & 0.1556 & 0.8694 \\
\midrule
\rowcolor[rgb]{ .886,  .937,  .855}
\textbf{StereoAdapter-2 (Ours)}   & \textbf{0.1023} & \textbf{0.5843} & \textbf{1.7164} & \textbf{0.1354} & \textbf{0.9256} \\
\bottomrule
\end{tabular}}
\end{table}

%% file: tables/ablation_model.tex
\begin{table}[t]
\centering
\caption{Model ablation of StereoAdapter-2, evaluating the effects of different design components, including the Depth Anything 3 encoder, monocular disparity initialization, context encoder, and update module.}

\label{tab:ablation_model}
\setlength{\tabcolsep}{4pt}
\renewcommand{\arraystretch}{1.1}
\small
\begin{tabularx}{\columnwidth}{@{} c c c c r r @{}}
\toprule
\makecell[c]{DA3\\Encoder}
& \makecell[c]{Mono Disp.\\Init.}
& \makecell[c]{Context\\Encoder}
& \makecell[c]{Update\\Module}
& \textbf{REL$\downarrow$}
& \textbf{RMSE$\downarrow$} \\
\midrule
        &        & \checkmark & ConvGRU  & 0.0516 & 2.82 \\
\checkmark &        & \checkmark & ConvGRU  & 0.0482 & 2.64 \\
\checkmark &        & \checkmark & ConvSS2D & 0.0449 & 2.46 \\
\checkmark &        &            & ConvSS2D & 0.0463 & 2.54 \\
\rowcolor[rgb]{.886,.937,.855}
\checkmark & \checkmark &            & ConvSS2D &
\textbf{0.0440} &
\textbf{2.40} \\
\bottomrule
\end{tabularx}
\vspace{-1em}
\end{table}

%% file: tables/ablation_hyperparameters.tex
\begin{table}[t]
\centering
\caption{Ablation on training hyperparameters.}
\label{tab:ablation_training}
\setlength{\tabcolsep}{6pt}
\renewcommand{\arraystretch}{1.0}
\small
\begin{tabular}{c c c r r}
\toprule
\textbf{Batch Size} 
& \textbf{Learning Rate} 
& \textbf{Train Iters} 
& \textbf{REL$\downarrow$} 
& \textbf{RMSE$\downarrow$} \\
\midrule
4  & $1\times10^{-4}$ & 16 & 0.0461 & 2.53 \\
4  & $2\times10^{-4}$ & 16 & 0.0489 & 2.68 \\
8  & $1\times10^{-4}$ & 16 & 0.0453 & 2.47 \\
8  & $2\times10^{-4}$ & 16 & 0.0476 & 2.59 \\
\rowcolor[rgb]{.886,.937,.855}
8  & $1\times10^{-4}$ & 22 & \textbf{0.0440} & \textbf{2.40} \\
\bottomrule
\end{tabular}
\vspace{-2em}
\end{table}

%% file: tables/ablation_ss2d.tex
\begin{table}[t]
\centering
\caption{Ablation study on ConvSS2D SSM hyperparameters under FP32 precision, analyzing the effects of the state dimension $d_{\text{state}}$ and SSM ratio on model accuracy and efficiency.}
\label{tab:ablation_ssm}
\setlength{\tabcolsep}{3pt}
\renewcommand{\arraystretch}{1.15}
\small
\begin{tabular}{c c c c c r r}
\toprule
\textbf{d\_state}
& \textbf{ssm\_ratio}
& \textbf{Params}
& \textbf{FLOPs}
& \textbf{TP.}
& \textbf{REL$\downarrow$}
& \textbf{RMSE$\downarrow$} \\

&
&
\textbf{(M)}
& \textbf{(G)}
& \textbf{(img/s)}
&
& \\

\midrule

1  & 1.0 & 20.40 & 843.33 & 5.26 & 0.0445 & 2.42 \\
\rowcolor[rgb]{.886,.937,.855}
4  & 1.0 & 20.42 & 845.81 & 5.22 & {0.0440} & {2.40} \\
16 & 1.0 & 20.47 & 855.72 & 4.71 & \textbf{0.0438} & \textbf{2.38} \\
16 & 1.5 & 20.60 & 886.17 & 4.62 & 0.4430 & 2.43 \\
16 & 2.0 & 20.73 & 916.62 & 4.26 & 0.4551& 2.45 \\
\bottomrule

\end{tabular}
\end{table}

%% file: tables/ablation_scan.tex
\begin{table}[t]
\centering
\caption{Ablation on SS2D scanning patterns.}
\label{tab:ablation_scanning}
\setlength{\tabcolsep}{3pt}
\renewcommand{\arraystretch}{1.15}
\small
\begin{tabular}{l c c c r r}
\toprule
\textbf{Scanning Pattern}
& \textbf{Params} 
& \textbf{FLOPs} 
& \textbf{TP.} 
& \textbf{REL$\downarrow$}
& \textbf{RMSE$\downarrow$} \\

& \textbf{(M)}
& \textbf{(G)}
& \textbf{(img/s)}
& 
& \\

\midrule
Unidi-Scan
& 20.47 & 855.72 & 4.87 &  0.0459 & 2.46 \\

Bidi-Scan
& 20.47 & 855.72 & 4.86 &  0.0453 & 2.42 \\

\rowcolor[rgb]{.886,.937,.855}
Cross-Scan
& 20.47 & 855.72 & 4.74 &  \textbf{0.0440} & \textbf{2.40} \\
\bottomrule
\end{tabular}
\end{table}

%% file: tables/inference_time.tex
\begin{table}[!h]
\centering
\caption{Average per-frame \textbf{inference latency (ms)} on Jetson Orin NX @ $640{\times}360$, batch size=1.}
\label{tab:orin_nx_latency}
\vspace{2pt}
\setlength{\tabcolsep}{6pt}
\renewcommand{\arraystretch}{1.05}
\begin{tabularx}{\columnwidth}{X c c}
\toprule
\textbf{Method} & \textbf{Params (M)} & \textbf{On-board (ms)} \\
\midrule
FoundationStereo \cite{FoundationStereo} & 375 & 1933 \\
Stereo Anywhere \cite{Stereoanywhere}   & 347 & 1524 \\
MGStereo \cite{squid}                   & 347 & 1631 \\
StereoAdapter \cite{StereoAdapter}      & 202 & 1285 \\
\rowcolor[rgb]{.886,.937,.855}
StereoAdapter-2 (Ours)                  & \textbf{103} & \textbf{1102} \\
\bottomrule
\end{tabularx}
\vspace{-1em}
\end{table}

%% file: appendix.tex
\appendix

\section{Appendix}

\begin{figure*}[b]

  \centering
  \includegraphics[width=\textwidth]{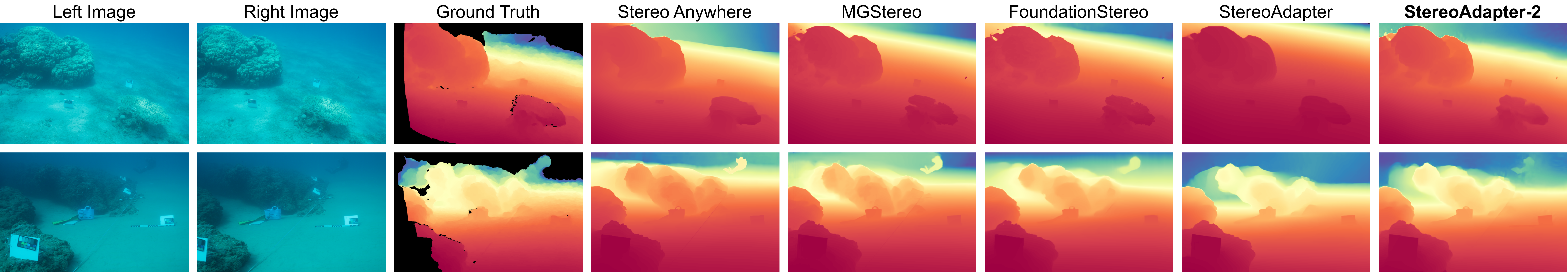}
  \captionof{figure}{Qualitative results of zero-shot stereo depth estimation for different models on the SQUID dataset.}
  \label{fig:appendix_squid_visualization}
  \vspace{0.3cm}
  \includegraphics[width=\textwidth,trim=0 18mm 0 18mm,clip]{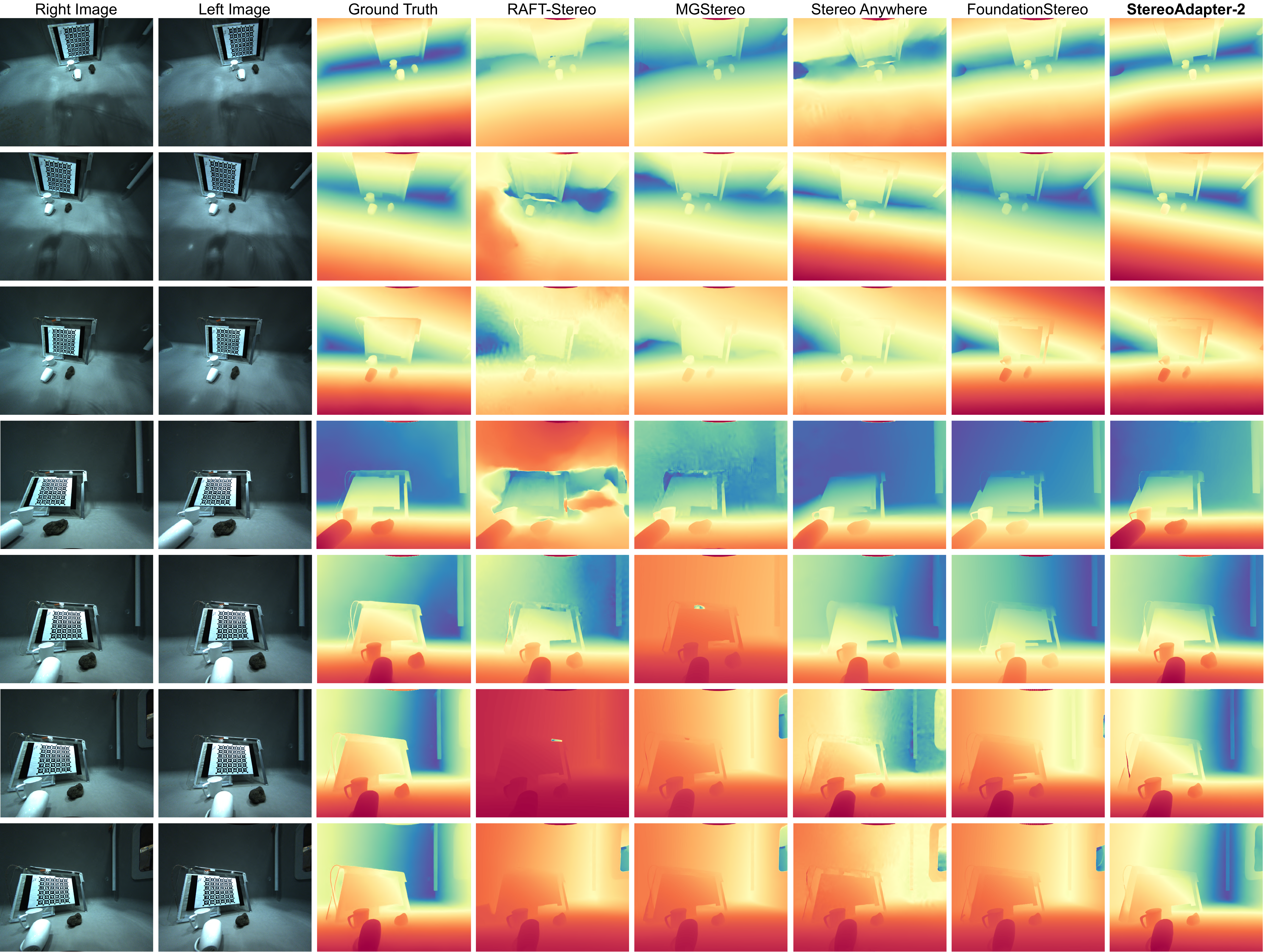}
  \captionof{figure}{Qualitative results of zero-shot stereo depth estimation for different models on the robot platform.}
  \label{fig:appendix_real_world_visualization}
\end{figure*}

\begin{figure*}[t]
  \centering
  \includegraphics[width=\textwidth]{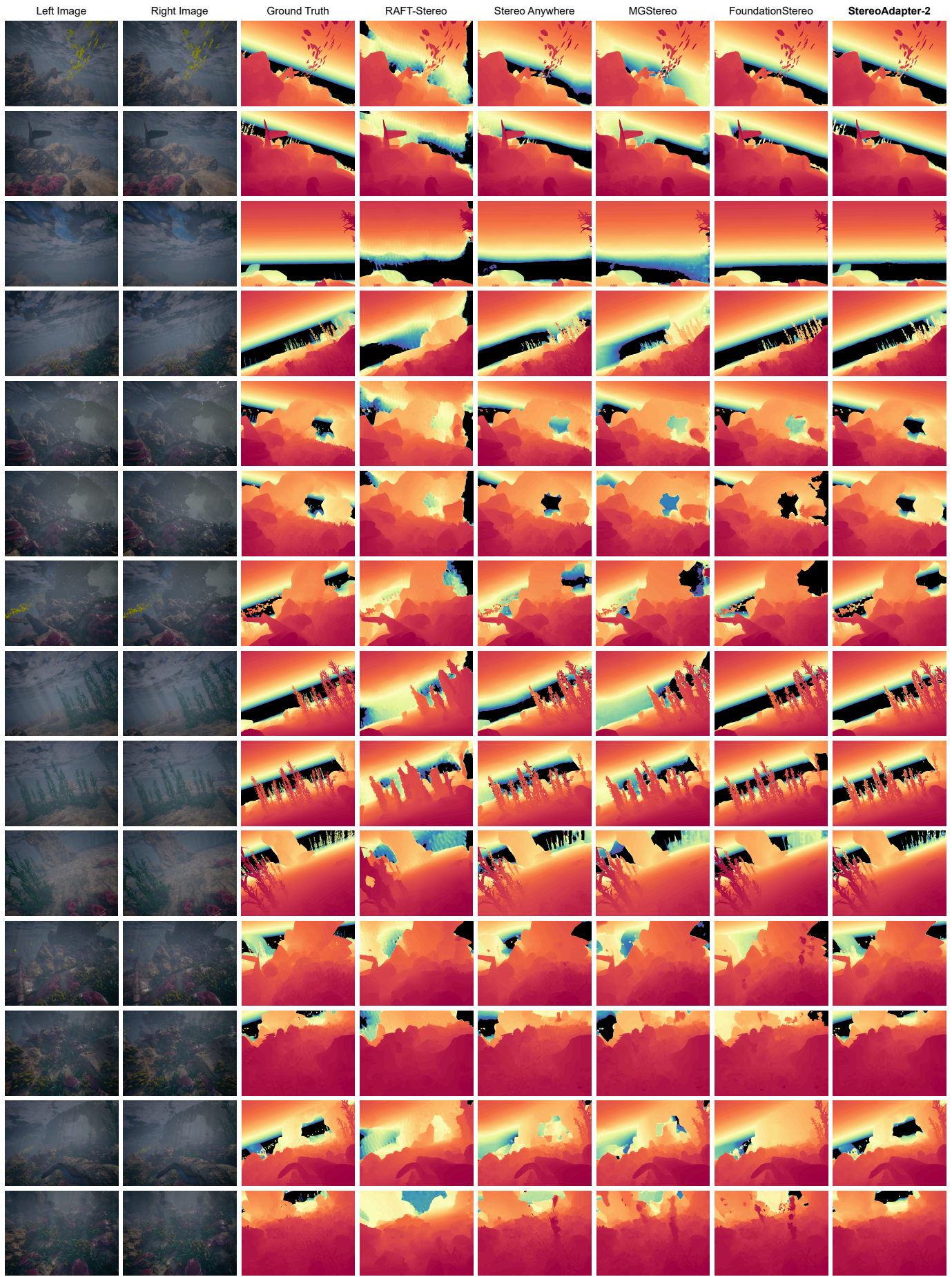}
  \caption{Qualitative results of zero-shot stereo depth estimation for different models on the Tartanair Ocean dataset}
  \label{fig:appendix_tartanair_visualization}
\end{figure*}

\begin{figure*}[t]
  \centering
  \includegraphics[width=\textwidth]{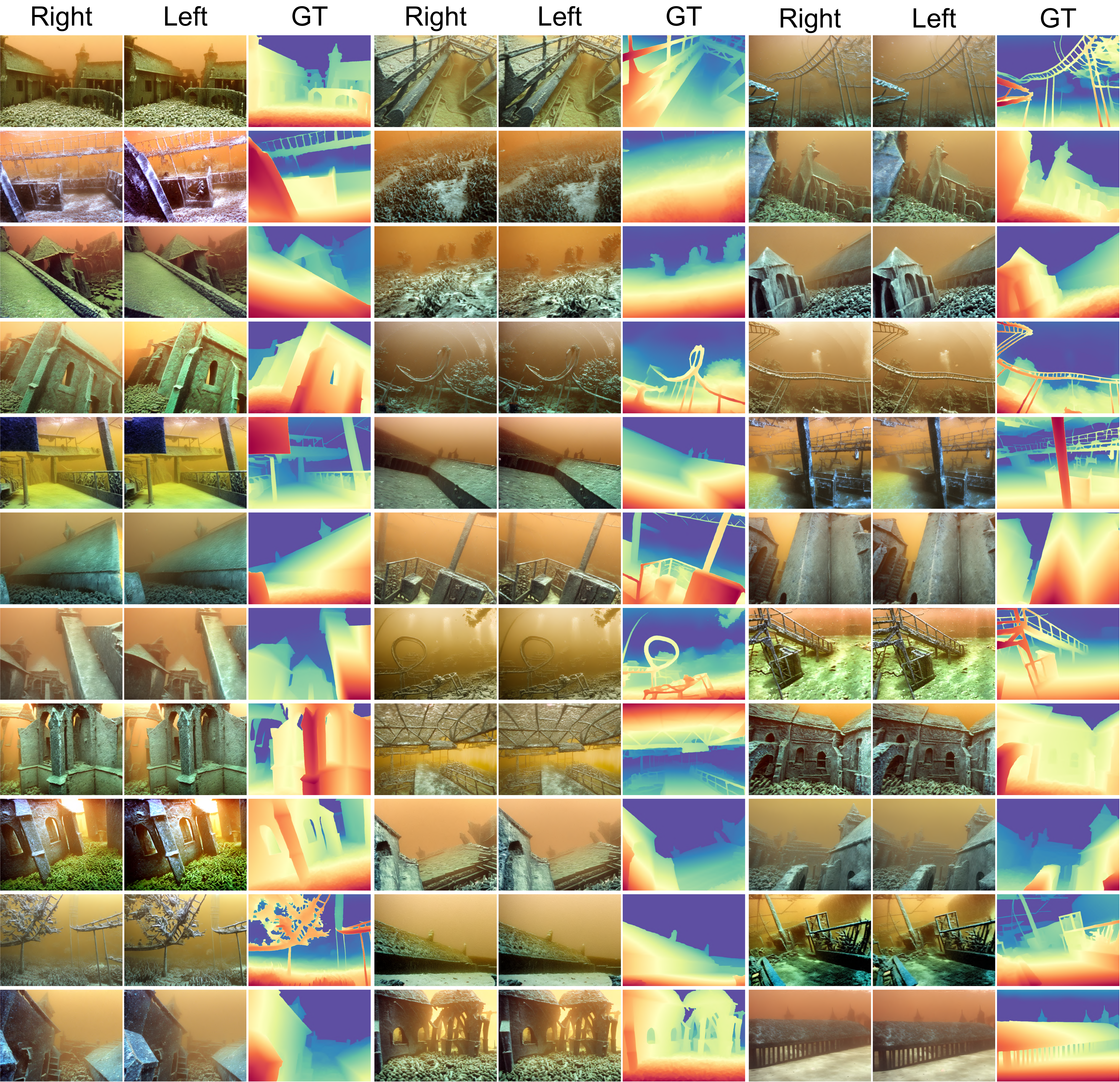}
  \caption{Visualization of UW-StereoDepth-80K}
  \label{fig:appendix_dataset_visualization}
\end{figure*}

\vspace*{0.5cm}
Figure \ref{fig:appendix_squid_visualization} shows qualitative comparisons of zero-shot stereo depth estimation on the SQUID dataset. Compared to existing methods, our approach produces more coherent disparity maps with clearer object boundaries and fewer artifacts in textureless and low-contrast underwater regions.
Figure \ref{fig:appendix_real_world_visualization} shows qualitative results obtained on a real-world robotic platform. The proposed method demonstrates stable and consistent depth predictions under real underwater conditions.
\newpage
\vspace*{0.5cm}
Figure \ref{fig:appendix_tartanair_visualization} shows qualitative zero-shot stereo depth estimation results on the TartanAir Ocean dataset.
Our method preserves fine-grained structural details and large-disparity regions more effectively, indicating strong generalization under diverse underwater appearances.
Figure \ref{fig:appendix_dataset_visualization} shows representative samples from the proposed UW-StereoDepth-80K dataset.
The dataset covers diverse underwater scenes, baselines, providing rich supervision for large-scale underwater stereo adaptation.


%% file: references.bib
@inproceedings{tartanair,
  title={Tartanair: A dataset to push the limits of visual slam. In 2020 IEEE},
  author={Wang, Wenshan and Zhu, Delong and Wang, Xiangwei and Hu, Yaoyu and Qiu, Yuheng and Wang, Chen and Hu, Yafei and Kapoor, Ashish and Scherer, Sebastian},
  booktitle={RSJ International Conference on Intelligent Robots and Systems (IROS)},
  pages={4909--4916}
}

@article{nvssolver,
  title={Nvs-solver: Video diffusion model as zero-shot novel view synthesizer},
  author={You, Meng and Zhu, Zhiyu and Liu, Hui and Hou, Junhui},
  journal={arXiv preprint arXiv:2405.15364},
  year={2024}
}

@inproceedings{Raft-stereo,
  title={Raft-stereo: Multilevel recurrent field transforms for stereo matching},
  author={Lipson, Lahav and Teed, Zachary and Deng, Jia},
  booktitle={2021 International Conference on 3D Vision (3DV)},
  pages={218--227},
  year={2021},
  organization={IEEE}
}

@article{da2,
  title={Depth anything v2},
  author={Yang, Lihe and Kang, Bingyi and Huang, Zilong and Zhao, Zhen and Xu, Xiaogang and Feng, Jiashi and Zhao, Hengshuang},
  journal={Advances in Neural Information Processing Systems},
  volume={37},
  pages={21875--21911},
  year={2024}
}

@article{vmamba,
  title={Vmamba: Visual state space model},
  author={Liu, Yue and Tian, Yunjie and Zhao, Yuzhong and Yu, Hongtian and Xie, Lingxi and Wang, Yaowei and Ye, Qixiang and Jiao, Jianbin and Liu, Yunfan},
  journal={Advances in neural information processing systems},
  volume={37},
  pages={103031--103063},
  year={2024}
}

@inproceedings{aanet,
  title={Aanet: Adaptive aggregation network for efficient stereo matching},
  author={Xu, Haofei and Zhang, Juyong},
  booktitle={Proceedings of the IEEE/CVF conference on computer vision and pattern recognition},
  pages={1959--1968},
  year={2020}
}

@inproceedings{waveletstereo,
  title={Waveletstereo: Learning wavelet coefficients of disparity map in stereo matching},
  author={Yang, Menglong and Wu, Fangrui and Li, Wei},
  booktitle={Proceedings of the IEEE/CVF conference on computer vision and pattern recognition},
  pages={12885--12894},
  year={2020}
}

@inproceedings{cfnet,
  title={Cfnet: Cascade and fused cost volume for robust stereo matching},
  author={Shen, Zhelun and Dai, Yuchao and Rao, Zhibo},
  booktitle={Proceedings of the IEEE/CVF conference on computer vision and pattern recognition},
  pages={13906--13915},
  year={2021}
}

@inproceedings{uasnet,
  title={Uasnet: Uncertainty adaptive sampling network for deep stereo matching},
  author={Mao, Yamin and Liu, Zhihua and Li, Weiming and Dai, Yuchao and Wang, Qiang and Kim, Yun-Tae and Lee, Hong-Seok},
  booktitle={Proceedings of the IEEE/CVF International Conference on Computer Vision},
  pages={6311--6319},
  year={2021}
}

@inproceedings{sednet,
  title={Learning the distribution of errors in stereo matching for joint disparity and uncertainty estimation},
  author={Chen, Liyan and Wang, Weihan and Mordohai, Philippos},
  booktitle={Proceedings of the IEEE/CVF Conference on Computer Vision and Pattern Recognition},
  pages={17235--17244},
  year={2023}
}

@inproceedings{pcwnet,
  title={Pcw-net: Pyramid combination and warping cost volume for stereo matching},
  author={Shen, Zhelun and Dai, Yuchao and Song, Xibin and Rao, Zhibo and Zhou, Dingfu and Zhang, Liangjun},
  booktitle={European conference on computer vision},
  pages={280--297},
  year={2022},
  organization={Springer}
}

@inproceedings{iResNet,
  title={Learning for disparity estimation through feature constancy},
  author={Liang, Zhengfa and Feng, Yiliu and Guo, Yulan and Liu, Hengzhu and Chen, Wei and Qiao, Linbo and Zhou, Li and Zhang, Jianfeng},
  booktitle={Proceedings of the IEEE conference on computer vision and pattern recognition},
  pages={2811--2820},
  year={2018}
}

@inproceedings{edgestereo,
  title={Edgestereo: A context integrated residual pyramid network for stereo matching},
  author={Song, Xiao and Zhao, Xu and Hu, Hanwen and Fang, Liangji},
  booktitle={Asian conference on computer vision},
  pages={20--35},
  year={2018},
  organization={Springer}
}

@inproceedings{HD3,
  title={Hierarchical discrete distribution decomposition for match density estimation},
  author={Yin, Zhichao and Darrell, Trevor and Yu, Fisher},
  booktitle={Proceedings of the IEEE/CVF conference on computer vision and pattern recognition},
  pages={6044--6053},
  year={2019}
}

@inproceedings{CoEx,
  title={Correlate-and-excite: Real-time stereo matching via guided cost volume excitation},
  author={Bangunharcana, Antyanta and Cho, Jae Won and Lee, Seokju and Kweon, In So and Kim, Kyung-Soo and Kim, Soohyun},
  booktitle={2021 IEEE/RSJ International Conference on Intelligent Robots and Systems (IROS)},
  pages={3542--3548},
  year={2021},
  organization={IEEE}
}

@inproceedings{PSMNet,
  title={Pyramid stereo matching network},
  author={Chang, Jia-Ren and Chen, Yong-Sheng},
  booktitle={Proceedings of the IEEE conference on computer vision and pattern recognition},
  pages={5410--5418},
  year={2018}
}

@inproceedings{GwcNet,
  title={Group-wise correlation stereo network},
  author={Guo, Xiaoyang and Yang, Kai and Yang, Wukui and Wang, Xiaogang and Li, Hongsheng},
  booktitle={Proceedings of the IEEE/CVF conference on computer vision and pattern recognition},
  pages={3273--3282},
  year={2019}
}

@inproceedings{gcnet,
  title={End-to-end learning of geometry and context for deep stereo regression},
  author={Kendall, Alex and Martirosyan, Hayk and Dasgupta, Saumitro and Henry, Peter and Kennedy, Ryan and Bachrach, Abraham and Bry, Adam},
  booktitle={Proceedings of the IEEE international conference on computer vision},
  pages={66--75},
  year={2017}
}

@inproceedings{hsm,
  title={Hierarchical deep stereo matching on high-resolution images},
  author={Yang, Gengshan and Manela, Joshua and Happold, Michael and Ramanan, Deva},
  booktitle={Proceedings of the IEEE/CVF Conference on Computer Vision and Pattern Recognition},
  pages={5515--5524},
  year={2019}
}

@inproceedings{parameterized,
  title={Parameterized cost volume for stereo matching},
  author={Zeng, Jiaxi and Yao, Chengtang and Yu, Lidong and Wu, Yuwei and Jia, Yunde},
  booktitle={Proceedings of the IEEE/CVF International Conference on Computer Vision},
  pages={18347--18357},
  year={2023}
}

@inproceedings{Ga-net,
  title={Ga-net: Guided aggregation net for end-to-end stereo matching},
  author={Zhang, Feihu and Prisacariu, Victor and Yang, Ruigang and Torr, Philip HS},
  booktitle={Proceedings of the IEEE/CVF conference on computer vision and pattern recognition},
  pages={185--194},
  year={2019}
}

@inproceedings{Sgm-nets,
  title={Sgm-nets: Semi-global matching with neural networks},
  author={Seki, Akihito and Pollefeys, Marc},
  booktitle={Proceedings of the IEEE conference on computer vision and pattern recognition},
  pages={231--240},
  year={2017}
}

@inproceedings{spyropoulos2014learning,
  title={Learning to detect ground control points for improving the accuracy of stereo matching},
  author={Spyropoulos, Aristotle and Komodakis, Nikos and Mordohai, Philippos},
  booktitle={Proceedings of the IEEE conference on computer vision and pattern recognition},
  pages={1621--1628},
  year={2014}
}

@article{Ndr,
  title={Neural disparity refinement},
  author={Tosi, Fabio and Aleotti, Filippo and Ramirez, Pierluigi Zama and Poggi, Matteo and Salti, Samuele and Mattoccia, Stefano and Di Stefano, Luigi},
  journal={IEEE Transactions on Pattern Analysis and Machine Intelligence},
  volume={46},
  number={12},
  pages={8900--8917},
  year={2024},
  publisher={IEEE}
}

@inproceedings{csm,
  title={Computing the stereo matching cost with a convolutional neural network},
  author={Zbontar, Jure and LeCun, Yann},
  booktitle={Proceedings of the IEEE conference on computer vision and pattern recognition},
  pages={1592--1599},
  year={2015}
}

@article{MC-CNN-acrt,
  title={Stereo matching by training a convolutional neural network to compare image patches},
  author={{\v{Z}}bontar, Jure and LeCun, Yann},
  journal={Journal of Machine Learning Research},
  volume={17},
  number={65},
  pages={1--32},
  year={2016}
}

@inproceedings{Orstereo,
  title={Orstereo: Occlusion-aware recurrent stereo matching for 4k-resolution images},
  author={Hu, Yaoyu and Wang, Wenshan and Yu, Huai and Zhen, Weikun and Scherer, Sebastian},
  booktitle={2021 IEEE/RSJ International Conference on Intelligent Robots and Systems (IROS)},
  pages={5671--5678},
  year={2021},
  organization={IEEE}
}

@inproceedings{CREStereo,
  title={Practical stereo matching via cascaded recurrent network with adaptive correlation},
  author={Li, Jiankun and Wang, Peisen and Xiong, Pengfei and Cai, Tao and Yan, Ziwei and Yang, Lei and Liu, Jiangyu and Fan, Haoqiang and Liu, Shuaicheng},
  booktitle={Proceedings of the IEEE/CVF conference on computer vision and pattern recognition},
  pages={16263--16272},
  year={2022}
}

@inproceedings{Eai-stereo,
  title={Eai-stereo: Error aware iterative network for stereo matching},
  author={Zhao, Haoliang and Zhou, Huizhou and Zhang, Yongjun and Zhao, Yong and Yang, Yitong and Ouyang, Ting},
  booktitle={Proceedings of the Asian conference on computer vision},
  pages={315--332},
  year={2022}
}

@inproceedings{IGEV-Stereo,
  title={Iterative geometry encoding volume for stereo matching},
  author={Xu, Gangwei and Wang, Xianqi and Ding, Xiaohuan and Yang, Xin},
  booktitle={Proceedings of the IEEE/CVF conference on computer vision and pattern recognition},
  pages={21919--21928},
  year={2023}
}

@inproceedings{DLNR,
  title={High-frequency stereo matching network},
  author={Zhao, Haoliang and Zhou, Huizhou and Zhang, Yongjun and Chen, Jie and Yang, Yitong and Zhao, Yong},
  booktitle={Proceedings of the IEEE/CVF conference on computer vision and pattern recognition},
  pages={1327--1336},
  year={2023}
}

@inproceedings{CREStereo++,
  title={Uncertainty guided adaptive warping for robust and efficient stereo matching},
  author={Jing, Junpeng and Li, Jiankun and Xiong, Pengfei and Liu, Jiangyu and Liu, Shuaicheng and Guo, Yichen and Deng, Xin and Xu, Mai and Jiang, Lai and Sigal, Leonid},
  booktitle={Proceedings of the IEEE/CVF International Conference on Computer Vision},
  pages={3318--3327},
  year={2023}
}

@inproceedings{Selective-stereo,
  title={Selective-stereo: Adaptive frequency information selection for stereo matching},
  author={Wang, Xianqi and Xu, Gangwei and Jia, Hao and Yang, Xin},
  booktitle={Proceedings of the IEEE/CVF Conference on Computer Vision and Pattern Recognition},
  pages={19701--19710},
  year={2024}
}

@inproceedings{Any-stereo,
  title={Any-stereo: Arbitrary scale disparity estimation for iterative stereo matching},
  author={Liang, Zhaohuai and Li, Changhe},
  booktitle={Proceedings of the AAAI Conference on Artificial Intelligence},
  volume={38},
  number={4},
  pages={3333--3341},
  year={2024}
}

@InProceedings{XR-Stereo,
    author    = {Cheng, Ziang and Yang, Jiayu and Li, Hongdong},
    title     = {Stereo Matching in Time: 100+ FPS Video Stereo Matching for Extended Reality},
    booktitle = {Proceedings of the IEEE/CVF Winter Conference on Applications of Computer Vision (WACV)},
    month     = {January},
    year      = {2024},
    pages     = {8719-8728}
}

@article{Mc-stereo,
  title={Mc-stereo: Multi-peak lookup and cascade search range for stereo matching},
  author={Feng, Miaojie and Cheng, Junda and Jia, Hao and Liu, Longliang and Xu, Gangwei and Hu, Qingyong and Yang, Xin},
  journal={arXiv preprint arXiv:2311.02340},
  year={2023}
}

@inproceedings{Mocha-stereo,
  title={Mocha-stereo: Motif channel attention network for stereo matching},
  author={Chen, Ziyang and Long, Wei and Yao, He and Zhang, Yongjun and Wang, Bingshu and Qin, Yongbin and Wu, Jia},
  booktitle={Proceedings of the IEEE/CVF conference on computer vision and pattern recognition},
  pages={27768--27777},
  year={2024}
}

@inproceedings{ICGNet,
  title={Learning intra-view and cross-view geometric knowledge for stereo matching},
  author={Gong, Rui and Liu, Weide and Gu, Zaiwang and Yang, Xulei and Cheng, Jun},
  booktitle={Proceedings of the IEEE/CVF conference on computer vision and pattern recognition},
  pages={20752--20762},
  year={2024}
}

@article{Igev++,
  title={Igev++: Iterative multi-range geometry encoding volumes for stereo matching},
  author={Xu, Gangwei and Wang, Xianqi and Zhang, Zhaoxing and Cheng, Junda and Liao, Chunyuan and Yang, Xin},
  journal={IEEE Transactions on Pattern Analysis and Machine Intelligence},
  year={2025},
  publisher={IEEE}
}

@inproceedings{STTR,
  title={Revisiting stereo depth estimation from a sequence-to-sequence perspective with transformers},
  author={Li, Zhaoshuo and Liu, Xingtong and Drenkow, Nathan and Ding, Andy and Creighton, Francis X and Taylor, Russell H and Unberath, Mathias},
  booktitle={Proceedings of the IEEE/CVF international conference on computer vision},
  pages={6197--6206},
  year={2021}
}

@inproceedings{CSTR,
  title={Context-enhanced stereo transformer},
  author={Guo, Weiyu and Li, Zhaoshuo and Yang, Yongkui and Wang, Zheng and Taylor, Russell H and Unberath, Mathias and Yuille, Alan and Li, Yingwei},
  booktitle={European conference on computer vision},
  pages={263--279},
  year={2022},
  organization={Springer}
}

@inproceedings{Chitransformer,
  title={Chitransformer: Towards reliable stereo from cues},
  author={Su, Qing and Ji, Shihao},
  booktitle={Proceedings of the IEEE/CVF conference on computer vision and pattern recognition},
  pages={1939--1949},
  year={2022}
}

@article{dynamicstereo,
  author    = {Nikita Karaev and Ignacio Rocco and Benjamin Graham and Natalia Neverova and Andrea Vedaldi and Christian Rupprecht},
  title     = {DynamicStereo: Consistent Dynamic Depth from Stereo Videos},
  journal   = {CVPR},
  year      = {2023},
}

@inproceedings{gmflow,
  title={Gmflow: Learning optical flow via global matching},
  author={Xu, Haofei and Zhang, Jing and Cai, Jianfei and Rezatofighi, Hamid and Tao, Dacheng},
  booktitle={Proceedings of the IEEE/CVF conference on computer vision and pattern recognition},
  pages={8121--8130},
  year={2022}
}

@inproceedings{croco_v2,
  title={{CroCo v2: Improved Cross-view Completion Pre-training for Stereo Matching and Optical Flow}},
  author={Weinzaepfel, Philippe and Lucas, Thomas and Leroy, Vincent and Cabon, Yohann and Arora, Vaibhav and Br{\'e}gier, Romain and Csurka, Gabriela and Antsfeld, Leonid and Chidlovskii, Boris and Revaud, J{\'e}r{\^o}me}, 
  booktitle={ICCV},
  year={2023}
}

@article{ELFNet,
  title={ELFNet: Evidential Local-global Fusion for Stereo Matching},
  author={Lou, Jieming and Liu, Weide and Chen, Zhuo and Liu, Fayao and Cheng, Jun},
  journal={arXiv preprint arXiv:2308.00728},
  year={2023}
}

@inproceedings{goat,
  title={Global occlusion-aware transformer for robust stereo matching},
  author={Liu, Zihua and Li, Yizhou and Okutomi, Masatoshi},
  booktitle={Proceedings of the IEEE/CVF Winter Conference on Applications of Computer Vision},
  pages={3535--3544},
  year={2024}
}

@article{StereoAdapter,
  title={StereoAdapter: Adapting Stereo Depth Estimation to Underwater Scenes},
  author={Wu, Zhengri and Wang, Yiran and Wen, Yu and Zhang, Zeyu and Wu, Biao and Tang, Hao},
  journal={arXiv preprint arXiv:2509.16415},
  year={2025}
}

@article{attention,
  title={Attention is all you need},
  author={Vaswani, Ashish and Shazeer, Noam and Parmar, Niki and Uszkoreit, Jakob and Jones, Llion and Gomez, Aidan N and Kaiser, {\L}ukasz and Polosukhin, Illia},
  journal={Advances in neural information processing systems},
  volume={30},
  year={2017}
}

@inproceedings{revised_underwater,
  title={A revised underwater image formation model},
  author={Akkaynak, Derya and Treibitz, Tali},
  booktitle={Proceedings of the IEEE conference on computer vision and pattern recognition},
  pages={6723--6732},
  year={2018}
}

@mastersthesis{flsea,
  title={Flsea: Underwater visual-inertial and stereo-vision forward-looking datasets},
  author={Randall, Yelena},
  year={2023},
  school={University of Haifa (Israel)}
}

@article{uwstereo,
  title={UWStereo: A Large Synthetic Dataset for Underwater Stereo Matching},
  author={Lv, Qingxuan and Dong, Junyu and Li, Yuezun and Chen, Sheng and Yu, Hui and Zhang, Shu and Wang, Wenhan},
  journal={IEEE Transactions on Circuits and Systems for Video Technology},
  year={2025},
  publisher={IEEE}
}

@article{UWNet,
  title={Reliable and Effective Stereo Matching for Underwater Scenes},
  author={Zhu, Lvwei and Gao, Ying and Zhang, Jiankai and Li, Yongqing and Li, Xueying},
  journal={Remote Sensing},
  volume={16},
  number={23},
  pages={4570},
  year={2024},
  publisher={MDPI}
}

@article{s4,
  title={Efficiently modeling long sequences with structured state spaces},
  author={Gu, Albert and Goel, Karan and R{\'e}, Christopher},
  journal={arXiv preprint arXiv:2111.00396},
  year={2021}
}

@article{mamba,
  title={Mamba: Linear-Time Sequence Modeling with Selective State Spaces},
  author={Gu, Albert and Dao, Tri},
  journal={arXiv preprint arXiv:2312.00752},
  year={2023}
}

@inproceedings{mamba2,
  title={Transformers are {SSM}s: Generalized Models and Efficient Algorithms Through Structured State Space Duality},
  author={Dao, Tri and Gu, Albert},
  booktitle={International Conference on Machine Learning (ICML)},
  year={2024}
}

@article{vim,
  title={Vision mamba: Efficient visual representation learning with bidirectional state space model},
  author={Zhu, Lianghui and Liao, Bencheng and Zhang, Qian and Wang, Xinlong and Liu, Wenyu and Wang, Xinggang},
  journal={arXiv preprint arXiv:2401.09417},
  year={2024}
}

@inproceedings{efficientvmamba,
  title={Efficientvmamba: Atrous selective scan for light weight visual mamba},
  author={Pei, Xiaohuan and Huang, Tao and Xu, Chang},
  booktitle={Proceedings of the AAAI Conference on Artificial Intelligence},
  volume={39},
  number={6},
  pages={6443--6451},
  year={2025}
}

@inproceedings{Squeeze-and-excitation,
  title={Squeeze-and-excitation networks},
  author={Hu, Jie and Shen, Li and Sun, Gang},
  booktitle={Proceedings of the IEEE conference on computer vision and pattern recognition},
  pages={7132--7141},
  year={2018}
}

@article{localmamba,
  title={LocalMamba: Visual State Space Model with Windowed Selective Scan},
  author={Huang, Tao and Pei, Xiaohuan and You, Shan and Wang, Fei and Qian, Chen and Xu, Chang},
  journal={arXiv preprint arXiv:2403.09338},
  year={2024}
}

@article{plainmamba,
  title={Plainmamba: Improving non-hierarchical mamba in visual recognition},
  author={Yang, Chenhongyi and Chen, Zehui and Espinosa, Miguel and Ericsson, Linus and Wang, Zhenyu and Liu, Jiaming and Crowley, Elliot J},
  journal={arXiv preprint arXiv:2403.17695},
  year={2024}
}

@inproceedings{Mamba-nd,
  title={Mamba-nd: Selective state space modeling for multi-dimensional data},
  author={Li, Shufan and Singh, Harkanwar and Grover, Aditya},
  booktitle={European Conference on Computer Vision},
  pages={75--92},
  year={2024},
  organization={Springer}
}

@article{Simba,
  title={Simba: Simplified mamba-based architecture for vision and multivariate time series},
  author={Patro, Badri N and Agneeswaran, Vijay S},
  journal={arXiv preprint arXiv:2403.15360},
  year={2024}
}

@article{vit,
  title={An image is worth 16x16 words: Transformers for image recognition at scale},
  author={Dosovitskiy, Alexey},
  journal={arXiv preprint arXiv:2010.11929},
  year={2020}
}

@article{Vision_mamba,
  title={Vision mamba: A comprehensive survey and taxonomy},
  author={Liu, Xiao and Zhang, Chenxu and Huang, Fuxiang and Xia, Shuyin and Wang, Guoyin and Zhang, Lei},
  journal={IEEE Transactions on Neural Networks and Learning Systems},
  year={2025},
  publisher={IEEE}
}

@article{combining,
  title={Combining recurrent, convolutional, and continuous-time models with linear state space layers},
  author={Gu, Albert and Johnson, Isys and Goel, Karan and Saab, Khaled and Dao, Tri and Rudra, Atri and R{\'e}, Christopher},
  journal={Advances in neural information processing systems},
  volume={34},
  pages={572--585},
  year={2021}
}

@article{diagonal,
  title={Diagonal state spaces are as effective as structured state spaces},
  author={Gupta, Ankit and Gu, Albert and Berant, Jonathan},
  journal={Advances in neural information processing systems},
  volume={35},
  pages={22982--22994},
  year={2022}
}

@article{long_sequence_modeling,
  title={What makes convolutional models great on long sequence modeling?},
  author={Li, Yuhong and Cai, Tianle and Zhang, Yi and Chen, Deming and Dey, Debadeepta},
  journal={arXiv preprint arXiv:2210.09298},
  year={2022}
}

@article{da3,
  title={Depth Anything 3: Recovering the visual space from any views},
  author={Haotong Lin and Sili Chen and Jun Hao Liew and Donny Y. Chen and Zhenyu Li and Guang Shi and Jiashi Feng and Bingyi Kang},
  journal={arXiv preprint arXiv:2511.10647},
  year={2025}
}

@article{ren2026anydepth,
  title={AnyDepth: Depth Estimation Made Easy},
  author={Ren, Zeyu and Zhang, Zeyu and Li, Wukai and Liu, Qingxiang and Tang, Hao},
  journal={arXiv e-prints},
  pages={arXiv--2601},
  year={2026}
}

@article{LEAStereo,
  title={Hierarchical Neural Architecture Search for Deep Stereo Matching},
  author={Cheng, Xuelian and Zhong, Yiran and Harandi, Mehrtash and Dai, Yuchao and Chang, Xiaojun and Li, Hongdong and Drummond, Tom and Ge, Zongyuan},
  journal={Advances in Neural Information Processing Systems},
  volume={33},
  year={2020}
}

@article{ACVNet,
  title={Accurate and efficient stereo matching via attention concatenation volume},
  author={Xu, Gangwei and Wang, Yun and Cheng, Junda and Tang, Jinhui and Yang, Xin},
  journal={IEEE Transactions on Pattern Analysis and Machine Intelligence},
  year={2023},
  publisher={IEEE}
}

@article{tiodepth,
  title={Two-in-One Depth: Bridging the Gap Between Monocular and Binocular Self-supervised Depth Estimation},
  author={Zhou, Zhengming and Dong, Qiulei},
  journal={arXiv preprint arXiv:2309.00933},
  year={2023}
}

@article{FoundationStereo,
  title={FoundationStereo: Zero-Shot Stereo Matching},
  author={Bowen Wen and Matthew Trepte and Joseph Aribido and Jan Kautz and Orazio Gallo and Stan Birchfield},
  journal={CVPR},
  year={2025}
}

@inproceedings{Stereoanywhere,
  title={Stereo anywhere: Robust zero-shot deep stereo matching even where either stereo or mono fail},
  author={Bartolomei, Luca and Tosi, Fabio and Poggi, Matteo and Mattoccia, Stefano},
  booktitle={Proceedings of the Computer Vision and Pattern Recognition Conference},
  pages={1013--1027},
  year={2025}
}

@article{gmstereo,
  title={Unifying flow, stereo and depth estimation},
  author={Xu, Haofei and Zhang, Jing and Cai, Jianfei and Rezatofighi, Hamid and Yu, Fisher and Tao, Dacheng and Geiger, Andreas},
  journal={IEEE Transactions on Pattern Analysis and Machine Intelligence},
  volume={45},
  number={11},
  pages={13941--13958},
  year={2023},
  publisher={IEEE}
}

@inproceedings{atlantis,
  title={Atlantis: Enabling Underwater Depth Estimation with Stable Diffusion},
  author={Zhang, Fan and You, Shaodi and Li, Yu and Fu, Ying},
  booktitle={Proceedings of the IEEE/CVF Conference on Computer Vision and Pattern Recognition},
  pages={11852--11861},
  year={2024}
}

@inproceedings{squid,
title={Diving into Haze-Lines: Color Restoration of Underwater Images},
author={Berman, D. and Treibitz, T. and Avidan, S.},
booktitle={Proceedings of the British Machine Vision Conference},
publisher = {BMVA Press}, year={2017}, 
}

@inproceedings{mayer2016large,
  title={A Large Dataset to Train Convolutional Networks for Disparity, Optical Flow, and Scene Flow Estimation},
  author={Mayer, Nikolaus and Ilg, Eddy and H{\"a}usser, Philip and Fischer, Philipp and Cremers, Daniel and Dosovitskiy, Alexey and Brox, Thomas},
  booktitle={Proceedings of the IEEE Conference on Computer Vision and Pattern Recognition (CVPR)},
  pages={4040--4048},
  year={2016}
}

@inproceedings{li2022practical,
  title={Practical stereo matching via cascaded recurrent network with adaptive correlation},
  author={Li, Jiankun and Wang, Peisen and Xiong, Pengfei and Cai, Tao and Yan, Ziwei and Yang, Lei and Liu, Jiangyu and Fan, Haoqiang and Liu, Shuaicheng},
  booktitle={Proceedings of the IEEE/CVF Conference on Computer Vision and Pattern Recognition},
  pages={16263--16272},
  year={2022}
}

@article{Bao2020InStereo2KAL,
  title={InStereo2K: a large real dataset for stereo matching in indoor scenes},
  author={Wei Bao and Wen Wang and Yuhua Xu and Yulan Guo and Siyu Hong and Xiaohu Zhang},
  journal={Science China Information Sciences},
  year={2020},
  volume={63},
  url={https://api.semanticscholar.org/CorpusID:221110870}
}

@inproceedings{schoeps2017cvpr,
  author = {Thomas Sch\"ops and Johannes L. Sch\"onberger and Silvano Galliani and Torsten Sattler and Konrad Schindler and Marc Pollefeys and Andreas Geiger},
  title = {A Multi-View Stereo Benchmark with High-Resolution Images and Multi-Camera Videos},
  booktitle = {Conference on Computer Vision and Pattern Recognition (CVPR)},
  year = {2017}
}

@inproceedings{butler2012naturalistic,
  title={A Naturalistic Open Source Movie for Optical Flow Evaluation},
  author={Butler, Daniel J and Wulff, Jonas and Stanley, Garrett B and Black, Michael J},
  booktitle={European Conference on Computer Vision (ECCV)},
  pages={611--625},
  year={2012}
}

@article{berman2020underwater,
  title={Underwater Single Image Color Restoration Using Haze-Lines and a New Quantitative Dataset},
  author={Berman, Dana and Levy, Deborah and Avidan, Shai and Treibitz, Tali},
  journal={IEEE Transactions on Pattern Analysis and Machine Intelligence},
  volume={43},
  number={8},
  pages={2822--2837},
  year={2020}
}
